\documentclass[notitlepage, 11pt]{article}
\usepackage[left=1in, right=1in, top=1in, bottom=1in]{geometry}

\usepackage{titling}
\usepackage[USenglish]{babel}
\usepackage{xcolor}
\definecolor{darkblue}{rgb}{0.0, 0.0, 0.55}
\definecolor{darkred}{RGB}{139,0,0}
\definecolor{darkgreen}{RGB}{0,100,0}
\definecolor{darkpurple}{RGB}{85,26,139}
\definecolor{lightpurple}{RGB}{106,90,205}
\usepackage[colorlinks=true,linkcolor=darkpurple,citecolor=darkpurple,urlcolor=darkpurple]{hyperref}
\addto\captionsUSenglish{%
}
\addto\extrasUSenglish{%
}

\usepackage{authblk}
\pretitle{\centering\Large}
\posttitle{\par}
\predate{}
\postdate{}

\usepackage{natbib}
\setcitestyle{open={(},close={)}} 

\usepackage{graphicx}
\usepackage{comment}

\usepackage{booktabs}

\usepackage{wrapfig}

\usepackage{lipsum}

\usepackage{float}

\usepackage{alex-macros}
\usepackage{bm}

\RequirePackage[T1]{fontenc}
\RequirePackage[tt=false, type1=true]{libertine}
\RequirePackage[varqu]{zi4}

\newcommand*{\memory}{{m}}
\newcommand*{\dkey}{{D_k}}
\newcommand*{\dphi}{{D_\phi}}
\newcommand*{\dval}{{D_v}}
\newcommand*{\linearfunctions}{{\Mcal_{\textsc{linear}}}}

\usepackage{etoolbox}
\usepackage{appendix}
\makeatletter
\patchcmd{\hyper@makecurrent}{%
    \ifx\Hy@param\Hy@chapterstring
        \let\Hy@param\Hy@chapapp
    \fi
}{%
    \iftoggle{inappendix}{%
        \@checkappendixparam{chapter}%
        \@checkappendixparam{section}%
        \@checkappendixparam{subsection}%
        \@checkappendixparam{subsubsection}%
        \@checkappendixparam{paragraph}%
        \@checkappendixparam{subparagraph}%
    }{}%
}{}{\errmessage{failed to patch}}

\newcommand*{\@checkappendixparam}[1]{%
    \def\@checkappendixparamtmp{#1}%
    \ifx\Hy@param\@checkappendixparamtmp
        \let\Hy@param\Hy@appendixstring
    \fi
}
\makeatletter

\newtoggle{inappendix}
\togglefalse{inappendix}

\apptocmd{\appendix}{\toggletrue{inappendix}}{}{\errmessage{failed to patch}}
\apptocmd{\subappendices}{\toggletrue{inappendix}}{}{\errmessage{failed to patch}}

\title{Test-time regression: a unifying framework for designing sequence models with associative memory}

\author[$^1$]{Ke Alexander Wang\footnote{Correspondence to \email{alxwang@cs.stanford.edu}}}
\author[$^2$]{Jiaxin Shi\footnote{Now at Google Deepmind}}
\author[$^1$$^2$]{Emily B. Fox}
\affil[$^1$]{
  Department of Computer Science, Stanford University
}
\affil[$^2$]{
  Department of Statistics, Stanford University
}
\date{}

\begin{document}
\maketitle

\begin{abstract}

Sequence models lie at the heart of modern deep learning. However, rapid advancements have produced a diversity of seemingly unrelated architectures, such as Transformers and recurrent alternatives. In this paper, we introduce a unifying framework to understand and derive these sequence models, inspired by the empirical importance of associative recall, the capability to retrieve contextually relevant tokens. We formalize associative recall as a two-step process, memorization and retrieval, casting memorization as a regression problem. Layers that combine these two steps perform associative recall via ``test-time regression'' over its input tokens. Prominent layers, including linear attention, state-space models, fast-weight programmers, online learners, and softmax attention, arise as special cases defined by three design choices: the regression weights, the regressor function class, and the test-time optimization algorithm. Our approach clarifies how linear attention fails to capture inter-token correlations and offers a mathematical justification for the empirical effectiveness of query-key normalization in softmax attention. Further, it illuminates unexplored regions within the design space, which we use to derive novel higher-order generalizations of softmax attention. Beyond unification, our work bridges sequence modeling with classic regression methods, a field with extensive literature, paving the way for developing more powerful and theoretically principled architectures.

\end{abstract}

\section{Introduction}\label{sec:introduction}

Sequences play a vital role in modern machine learning by providing a powerful abstraction: any computational task can be viewed as transforming one sequence into another \citep{Sutskever2014SequenceSequenceLearning}.
This sequential perspective has spread across diverse domains, including natural language processing  \citep{Sutskever2014SequenceSequenceLearning,Devlin2019BERTPretrainingDeep,Brown2020LanguageModelsAre}, computer vision \citep{Dosovitskiy2021ImageWorth16x16,Bertasius2021SpaceTimeAttentionAll}, time series analysis \citep{Salinas2020DeepARProbabilisticForecasting,Gruver2023LargeLanguageModels,Ansari2024ChronosLearningLanguage}, and computational biology \citep{Jumper2021HighlyAccurateProtein,Zhou2015PredictingEffectsNoncoding,Nguyen2024SequenceModelingDesign}, highlighting the importance of building generically applicable sequence layers \citep{Vaswani2017AttentionAllYou}.

This development has produced a diversity of architectures, each with its own unique characteristics and performance trade-offs.
While these architectures have achieved considerable success, they have largely emerged through separate lines of investigation.
Such a fragmented and often empirically-driven approach to model development limits our ability to systematically understand and improve design choices.
Moreover, the idiosyncratic notations of each architecture obscures their underlying connections \citep{SashaRush2024ThereAre4}.
Given the wide variety of sequence models, a natural question to ask is whether there is an underlying principle that explains why some sequence models work better than others.

One empirical discovery that ties together disparate architectures is the strong correlation between an architecture's associative recall ability and its language modeling performance \citep{Olsson2022ContextLearningInduction,Arora2023ZoologyMeasuringImproving}.
Associative recall is the act of retrieving contextually relevant information based on an association with a query.
\citep{Arora2023ZoologyMeasuringImproving} gives the following example of how associative memory and retrieval can improve language modeling.
Consider the example sentence from \citet{Arora2023LanguageModelsEnable}: ``Hakuna Matata, it means no worries for the rest of your days. Hakuna \underline{\hspace{1cm}} ''.
To predict the next word, we can first memorize the previous occurrence of ``Hakuna'' and its associated value ``Matata''; the next time we encounter ``Hakuna'' again we can retrieve its associated value from memory as our prediction.
Notice that we can perform this task using only in-context information and we can make an accurate prediction even if we have never encountered this strange phrase before.
Indeed, transformer-based language models have been discovered to exhibit this kind of behavior via ``induction heads'' that emerge during training \citep{Olsson2022ContextLearningInduction}.

\begin{figure}
    \centering
    \includegraphics[width=\linewidth]{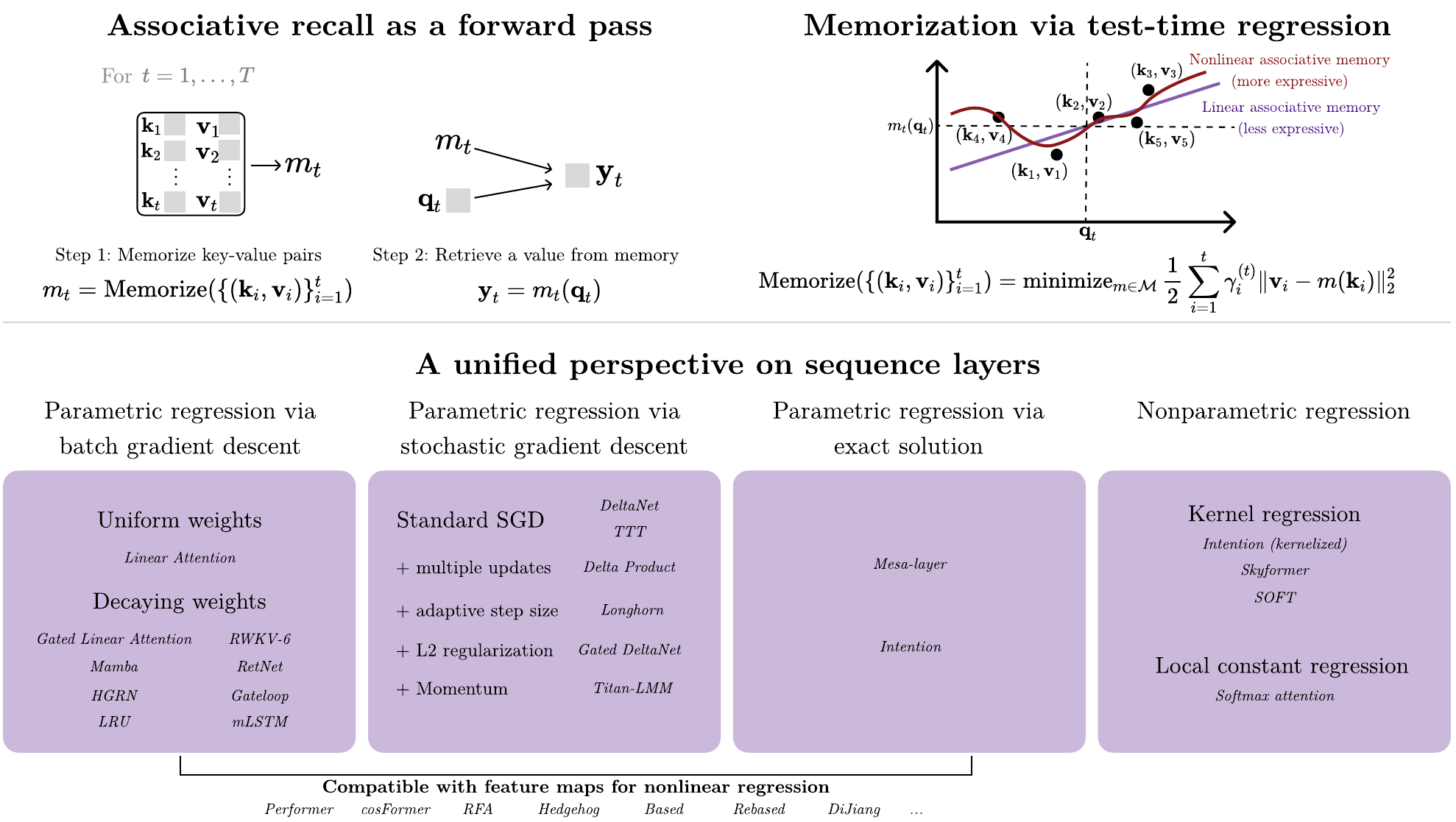}
    \caption{Our framework provides a systematic way to derive sequence models that can perform associative recall, following a two step process of memorization and retrieval. The memorization step can be formalized as a solving a regression problem at test-time. Our perspective results in a ``recipe'' for derving sequence layers by making three choices: the importance of each association $\{\gamma_i^{(t)}\}_{i=1}^t$, the regressor function class $\Mcal$, and the optimization algorithm. Parametric regression layers tend to have an efficient recurrence for updating the memory $\memory_t$, while non-parametric layers like softmax-attention do not. For simplicity, we discuss causal sequence models where prefix key-value pairs are memorized, but the same principles also apply to non-causal ones (e.g. generically masked attention).}
    \label{fig:landscape-diagram}
\end{figure}

Given the empirical importance of associative recall, \textbf{how can we systematically design neural network layers that can perform associative recall (AR)?}
In this paper, we introduce a simple but principled framework for deriving sequence layers \emph{designed to perform associative recall}, which we call ``test-time regression layers''.
From the ``Hakuna Matata'' example, we see that \textbf{associative recall has two-steps: memorization and retrieval}.
Our crucial observation is that we can implement the memorization step by solving a weighted regression problem.
We can then generate an output by applying the regressor to a cue/query token, retrieving the most relevant token from our associative memory.
\emph{Combining both memorization and retrieval into the forward pass of a sequence layer} results in a layer that performs regression over the input tokens, a procedure that we call ``test-time regression'', visualized in \autoref{fig:landscape-diagram}.
Our terminology reflects that the regressor is regenerated with each forward pass and depends only on the \emph{input tokens} rather than a fixed training dataset.

Since any regression method can be used, our framework provides a general recipe to derive a large class of sequence layers.
\textbf{Under our framework, an AR-based sequence layer is a mathematical consequence of choosing (1) the regression weights, (2) the regressor parameterization, and (3) the optimization algorithm for finding the regressor.}
In this paper, we consider a few choices of these three ``ingredients'', and show how to \emph{derive} many recently proposed classes of sequence layers, illustrating the generality of our framework.
Our derivations reveal that linear attention \citep{Katharopoulos2020TransformersAreRNNs}, its feature-mapped variants (e.g. \citep{Peng2020RandomFeatureAttention,Qin2021CosFormerRethinkingSoftmax,Kasai2021FinetuningPretrainedTransformers,Zhang2023HedgehogPorcupineExpressive,Aksenov2024LinearTransformersLearnable,Chen2024DiJiangEfficientLarge}), its gated variants \citep{Sun2023RetentiveNetworkSuccessor,Orvieto2023ResurrectingRecurrentNeural,Katsch2024GateLoopFullyDataControlled,De2024GriffinMixingGated,Qin2024HGRN2GatedLinear,Peng2024EagleFinchRWKV,Yang2024GatedLinearAttention,Beck2024XLSTMExtendedLong}, state-space model layers \citep{Gu2024MambaLinearTimeSequence,Dao2024TransformersAreSSMs}, fast-weights layers \citep{Schlag2021LinearTransformersAre,Yang2024GatedDeltaNetworks,Yang2024GatedDeltaNetworks}, online learning layers \citep{Liu2024LonghornStateSpace,Sun2024LearningLearnTest,Yang2024GatedDeltaNetworks,Behrouz2024TitansLearningMemorize}, and softmax self-attention \citep{Vaswani2017AttentionAllYou} \emph{are all test-time regression layers}, implicitly performing memorization followed by retrieval in their forward passes, despite being developed from different perspectives.
\autoref{fig:landscape-diagram} previews how existing architectures are instantiations of test-time regression layers within our framework.

Our derivations also lead to new understandings of existing sequence layers.
We show that layers based on linear attention underperform because they fail to account for the correlation between tokens.
We also show that query-key normalization, an important technique in stabilizing the training of large language models \citep{Dehghani2023ScalingVisionTransformers,Wortsman2023SmallscaleProxiesLargescale}, is mathematically necessary to ensure that softmax self-attention is a proper local constant regressor.
Finally, we propose a higher order generalization of softmax attention, motivated by local linear regression.

\paragraph{Outline of our paper.}
We start in \autoref{sec:existing-archs} by examining a few of the most prominent classes of sequence layers.
Although each class of layers was developed from distinct motivations, their similar computation pattern hints at a common unifying theme that underlies their effectiveness.
We then introduce our test-time regression framework in \autoref{sec:duality} to formalize the connection between associative memorization and regression.
Our perspective provides a systematic approach to designing sequence layers that produce its outputs via associative recall.
In \autoref{sec:deriving-architectures}, we show that indeed all of the aforementioned classes of sequence layers can be understood from a \emph{single unified perspective} using the principles of test-time regression and associative recall.
We demonstrate the generality of our framework by deriving these layers simply by varying how we minimize the regression objective.
\autoref{sec:experiments} empirically validates that test-time regression layers implicitly perform regression over its input tokens with a single forward pass.
Then, in \autoref{sec:key-construction} we examine how to construct effective key-value pairs for associative recall in next-token prediction tasks.
We discuss prior work related to associative recall and memory in \autoref{sec:related-works}.
We finish in \autoref{sec:discussion} by discussing the broader implications of viewing sequence models through the lens of regression, including future directions for new architectures.

\section{Background on existing sequence modeling layers}\label{sec:existing-archs}
The goal of sequence modeling is to transform a sequence of input tokens $\xbf_1, \ldots, \xbf_T$ into output tokens $\ybf_1, \ldots, \ybf_T$ \citep{Sutskever2014SequenceSequenceLearning}.
A common design pattern, first proposed by \citet{Vaswani2017AttentionAllYou} is to transform the input tokens into key-value pairs $(\kbf_1, \vbf_1), \ldots, (\kbf_T, \vbf_T) \in \reals^{\dkey}\times\reals^{\dval}$ and queries $\qbf_1, \ldots, \qbf_T \in \reals^\dkey$ before applying operations that ``mix'' the sequences together to produce the output tokens. %
Typically, each $\qbf_t, \kbf_t,$ and $\vbf_t$ is a linear transform of the corresponding input token $\xbf_t$, which can be thought of as different ``views'' of $\xbf_t$.
Most subsequent sequence layers have kept this pattern but propose other ways of ``mixing'' queries, keys, and values.
We begin by reviewing a few classes of existing sequence layers that follow this pattern before unifying them under our test-time regression framework.

\paragraph{Softmax attention.}
Given input tokens $\xbf_1, \ldots, \xbf_T$, and their query, key, value vectors, softmax attention \citep{Vaswani2017AttentionAllYou} computes each output token by
\begin{align}
    \ybf_t = \sum_{i=1}^t \vbf_i \frac{s(\kbf_i, \qbf_t)}{\sum_{j=1}^t s(\kbf_j, \qbf_t)} \label{eqn:softmax-attention}
\end{align}
for $t\in [T]$ where $s(\kbf, \qbf) = \exp(\kbf\transpose\qbf/\sqrt{\dkey})$.
Each output $\ybf_t$ is a weighted sum of values $\{\vbf_i\}_{i=1}^t$, weighted by the similarity between the query $\qbf_t$ and the keys $\{\kbf_i\}_{i=1}^t$ associated with the values.
This mechanism can be seen as a continuous-valued lookup table, using $\qbf_t$ to lookup an interpolated value for $\ybf_t$.
Though highly effective, self-attention requires $\bigo{T^2}$ time to autoregressively compute the entire length $T$ sequence even after caching the keys and values, which can be prohibitively slow for generating long sequences.

\paragraph{Linear attention.}
To address the computational challenges of self-attention, there has been a wealth of methods that seek to develop more efficient alternatives; see the survey by \citet{Tay2022EfficientTransformersSurvey}.
One particular class of more efficient sequence models has its origins in a recurrent variant of attention, known as linear attention \citep{Katharopoulos2020TransformersAreRNNs}.
To achieve this efficiency, linear attention uses the standard inner product $\kbf\transpose \qbf$, rather than using the exponentiated inner product $s(\kbf_i, \qbf_t)$, producing outputs of the form $\ybf_t \propto \sum_{i=1}^t \vbf_i\kbf_i\transpose\qbf_t$ up to a normalization.
This change allows all $T$ outputs to be recurrently computed in $\bigo{T}$ time, an asymptotic speedup, by updating a $\dval\times \dkey$ ``matrix-valued state'' $\Mbf_t = \sum_{i=1}^t\vbf_i\kbf_i\transpose$, following the recurrence
\begin{align}
    \Mbf_t = \Mbf_{t-1} + \vbf_t\kbf_t\transpose, \quad \ybf_t = \Mbf_t \qbf_t = \sum_{i=1}^t \vbf_i\kbf_i\transpose\qbf_t. \label{eqn:linear-attention-recurrence}
\end{align}
Each rank-1 outer product $\vbf_t\kbf_t\transpose$ can be thought of as writing a new key-value pair into memory.
Although \citet{Katharopoulos2020TransformersAreRNNs}'s original formulation of linear attention included a normalization like softmax attention, we drop the normalization here since later works found that this normalization method causes training instabilities \citep{Qin2022DevilLinearTransformer,Dao2024TransformersAreSSMs} .

To increase expressiveness, \citet{Katharopoulos2020TransformersAreRNNs} also proposed a variant of linear attention, inspired by the kernel trick \citep{Scholkopf2001LearningKernelsSupport}, that replaces the inner product $\kbf\transpose\qbf$ with an inner product in a feature space $\phi(\kbf)\transpose\phi(\qbf)$ where $\phi$ is a feature map $\phi:\dkey\to\dphi$.
This modification results in the recurrence
\begin{align}
    \Mbf_t = \Mbf_{t-1} + \vbf_t\phi(\kbf_t)\transpose, \quad \ybf_t = \Mbf_t \phi(\qbf_t) = \sum_{i=1}^t \vbf_i\phi(\kbf_i)\transpose\phi(\qbf_t).
\end{align}
\citet{Katharopoulos2020TransformersAreRNNs} used $\phi = 1 + \operatorname{ELU}$, but numerous alternative choices of feature maps have since been proposed, including the cosine function \citep{Qin2021CosFormerRethinkingSoftmax}, deterministic parameter-free projections \citep{Schlag2021LinearTransformersAre}, polynomial maps \citep{Zhang2023HedgehogPorcupineExpressive,Aksenov2024LinearTransformersLearnable,Arora2024SimpleLinearAttention}, random features \citep{Peng2020RandomFeatureAttention,Choromanski2020RethinkingAttentionPerformers,Chen2024DiJiangEfficientLarge}, and more.

\paragraph{State-space models and linear attention with forgetting.}
Since the recurrent state $\Mbf_t$ of linear attention has finite size ($\dval\times\dkey$ numbers), $\Mbf_t$ can only store a bounded amount of information.
In \autoref{eqn:linear-attention-recurrence}, $\Mbf_t$ weighs all key value pairs equally across timesteps, trying to compress all key-value pairs equally well into its finite-sized state.
As a result, the performance of linear attention can degrade as the sequence length increases.
To address this limitation, various works have proposed adding a forgetting factor $\gamma_t$ to allow $\Mbf_t$ to retain only the most recent information, decaying out key-value pairs over time.
This produces the recurrence
\begin{align}
    \Mbf_t = \gamma_t \Mbf_{t-1} + \vbf_t\kbf_t\transpose, \quad \ybf_t = \Mbf_t \qbf_t. \label{eqn:gated-linear-attention-recurrence}
\end{align}
At each step $t$, all previous values in $\Mbf_{t-1}$ are attenuated by a factor of $\gamma_t\in[0, 1]$ before the new value is stored.
When $\gamma_t$ depends on $\xbf_t$, this forgetting factor can be seen as a forgetting gate, akin to that of an LSTM \citep{Hochreiter1997LongShortTermMemory}.
This class of linear attention with forgetting includes RetNet \citep{Sun2023RetentiveNetworkSuccessor}, gated linear attention \citep{Yang2024GatedLinearAttention}, Gateloop \citep{Katsch2024GateLoopFullyDataControlled}, mLSTM \citep{Beck2024XLSTMExtendedLong}, LRU \citep{Orvieto2023ResurrectingRecurrentNeural}, HGRN \citep{Qin2024HGRN2GatedLinear}, and RWKV-6 \citep{Peng2024EagleFinchRWKV}.
See Table 1 of \citet{Yang2024GatedLinearAttention} for how each model implements $\gamma_t$.

The forgetting factor in linear attention is also equivalent to another class of sequence layers known as state-space models (SSMs), which includes Mamba \citep{Gu2024MambaLinearTimeSequence} and Mamba-2 \citep{Dao2024TransformersAreSSMs}.
State-space models originated from a line of work on optimal online orthogonal polynomial projections \citep{Gu2020HiPPORecurrentMemory,Gu2021EfficientlyModelingLong,Gu2022ParameterizationInitializationDiagonal} that enabled the architecture to retain information over long sequences.
However, many recent works \citep{Han2024DemystifyMambaVision,Trockman2024MimeticInitializationHelps,Ren2024CanMambaAlways}, including the Mamba-2 followup \citep{Dao2024TransformersAreSSMs}, have since shown that SSM layers are equivalent to linear attention with forgetting; the forgetting factor maps onto the time-varying step size and transition matrix used in SSMs.

\paragraph{Fast-weight layers and test-time training layers.}
An alternative to attenuating \emph{all} past information in $\Mbf_{t-1}$ is to forget only the values whose keys are most similar to the new key $\kbf_t$.
This was the motivation behind DeltaNet \citep{Schlag2021LinearTransformersAre}, which is a revival of classic fast weight layers \citep{Schmidhuber1992LearningControlFastWeight}.
DeltaNet follows a recurrence
\begin{align}
    \Mbf_{t} &= \Mbf_{t-1} - \underbrace{(\Mbf_{t-1}\kbf_t)}_{\vbf_{t}^{\text{old}}}\kbf_t\transpose + \underbrace{[\beta_t \vbf_t + (1-\beta_t) \overbrace{\Mbf_{t-1}\kbf_t}^{\vbf_t^{\text{old}}}]}_{\vbf_t^{\text{new}}}\kbf_t\transpose \label{eqn:fast-weights-interplation} \\
    &= \Mbf_{t-1}(\Ibf - \beta_t \kbf_t\kbf_t\transpose) + \beta_t \vbf_t\kbf_t\transpose \label{eqn:delta-rule}
\end{align}
where the first equation follows the notation of \citet{Schlag2021LinearTransformersAre} in defining $\beta_t\in [0, 1]$, $\vbf_t^{\text{old}}$, and $\vbf_t^{\text{new}}$.
Like previous layers, the output at step $t$ is also $\ybf_t = \Mbf_t \qbf_t.$
Comparing the recurrence of DeltaNet (\autoref{eqn:fast-weights-interplation}) to that of linear attention (\autoref{eqn:linear-attention-recurrence}), we can interpret the update rule as rewriting the value that was bound to key $\kbf_t$. 
The update rule deletes an old value $\vbf_t^{\text{old}}=\Mbf_{t-1}\kbf_t$ retrieved from the previous timestep's memory $\Mbf_{t-1}$ but using the new key $\kbf_t$.
The layer then updates the recurrent state with a new interpolated value $\vbf_t^{\text{new}}$, binding it to $\kbf_t$.

The iterates $\Mbf_t$ are also known as ``fast-weights'', originally proposed by \citet{Schmidhuber1992LearningControlFastWeight}.
These weights/iterates are ``fast'' because they change with each timestep, rapidly adapting to the input tokens, in contrast to the ``slow-weights'' in the rest of the neural network, such as the linear projections weights, which are only adjusted at training time.
The traditional ``slow-weights'' can be seen as the weights of a metalearner \citep{Clark2022MetaLearningFastWeight}, learning to use the ``fast-weights'' to learn from the input sequence \citep{Sun2024LearningLearnTest}.

\paragraph{Online learning layers.}
The recurrences we have reviewed so far all share a common high-level behavior: at each timestep $t$, the recurrent state must be adapted to incorporate information from the latest key-value pair $(\kbf_t, \vbf_t)$.
This property can be seen as making an online sequence of decisions, balancing between retaining historical information and adapting to new information.

Motivated by this sequential decision making perspective, \citet{Liu2024LonghornStateSpace} proposed to derive recurrences by minimizing an online sequence of objectives:
\begin{align}
    \Mbf_t = \argmin_\Mbf\ d(\Mbf_{t-1}, \Mbf) + \ell_t(\Mbf) \label{eqn:online-learning-objective}
\end{align}
where $d(\Mbf_{t-1}, \Mbf)$ is a measure of the discrepancy between $\Mbf$ and the previous iterate $\Mbf_{t-1}$, encouraging retention of older information, and $\ell_t$ is a loss function that encourages adapting to new information.
Once again, the output is computed by $\ybf_t = \Mbf_t\qbf_t$.
Recurrent layers derived via \autoref{eqn:online-learning-objective} are called ``online learning layers'' or ``online learners''.
Recent examples of such layers include Longhorn \citep{Liu2024LonghornStateSpace}, Gated DeltaNet\citep{Yang2024GatedDeltaNetworks}, and Titans \citep{Behrouz2024TitansLearningMemorize}, each of these layers was derived from specific choice of functions $d$ and $\ell_t$. See Table 2 from \citet{Yang2024GatedDeltaNetworks} for each layer's choice of $d$ and $\ell_t$.

\section{Test-time regression as a framework for designing sequence layers}\label{sec:duality}
Having discussed a number of existing sequence layers, each motivated by an atomic architectural modification, we now present a unified framework to clarify their relationship to each other.
Our approach is motivated by empirical findings that a model's associative recall ability strongly correlates with its performance and in-context learning capabilities \citep{Olsson2022ContextLearningInduction, Arora2023ZoologyMeasuringImproving}.
We formalize the two step process of memorization and retrieval, presenting a principled recipe for deriving sequence layers with mathematically grounded recall abilities.

\paragraph{Step 1: Memorization as regression.}
From our everyday experience, we already have an intuition for how associative memory should behave; for example, hearing a friend's name should trigger a mental impression of that friend.
This pairing of a ``cue'' and a ``response'' is known as an association.
To reflect terminology in deep learning \citep{Vaswani2017AttentionAllYou}, we will refer to cues as ``keys'' and responses as ``values''.
With this intuition, we can define a mathematical model for associative memory.
Given key-value pairs $(\kbf_1, \vbf_1), \ldots, (\kbf_t, \vbf_t) \in \reals^\dkey \times \reals^\dval$ to be memorized, an \emph{associative memory system} is a function $\memory: \reals^\dkey \to \reals^\dval$ such that $\memory(\kbf_i)\approx\vbf_i$ for $i=1, \ldots, t$ \citep{Kohonen1972CorrelationMatrixMemories,Kohonen1989SelfOrganizationAssociativeMemory}.
Such a mapping $\memory$ between two sets of related objects is also known as a hetero-associative memory in classic signal processing and neurocomputing literature \citep{Hinton1989ParallelModelsAssociative}.

Our main observation is that finding a memory map that satisfies these constraints reduces to finding a vector-valued regressor $\memory$.
We can then store a set of key-value associations into an associative memory system by solving the weighted regression problem
\begin{align}\label{eqn:general-least-squares}
    \text{Memorize associations:}\quad \quad \memory_t \approx \argmin_{\memory\in \Mcal}\ \frac{1}{2}\sum_{i=1}^t\gamma_i^{(t)}\norm{\vbf_i - \memory(\kbf_i)}_2^2,
\end{align}
where the importance of each association is controlled by adjusting the relative weights $\{\gamma_i^{(t)}\}_{i=1}^t$
Classic works on associative memory are restricted to linear memory maps, corresponding to linear regression  \citep[Chapter 6.6]{Kohonen1989SelfOrganizationAssociativeMemory}, but here we allow $\Mcal$ to be any function class.
We refer to this relationship between regression and associative memory as the ``regression-memory correspondence''.

\paragraph{Step 2: Memory retrieval as function application.}
Having derived a regression-based procedure to memorize a set of associations into $\memory$, we can now retrieve values from memory by applying $\memory$ to a new vector, which we call a ``query'' $\qbf$, producing output
\begin{align}
    \text{Memory retrieval:} \quad \quad \ybf_t = \memory_t(\qbf_t). 
\end{align}
The retrieval performance depends on how well \autoref{eqn:general-least-squares} is minimized.
When \autoref{eqn:general-least-squares} attains zero-loss, $\memory_t$ memorizes all associations perfectly, and we get perfect retrieval: $\memory_t(\kbf_i) = \vbf_i$ for $i= 1, \ldots, t$.
For certain function classes $\Mcal$, e.g. smooth functions, $\memory$ can also extrapolate outside of the existing set of keys.
In those cases, $\memory_t(\qbf)$ instead retrieves a value based on the proximity of $\qbf$ to the existing set of keys.

\paragraph{Sequence layers that memorize then retrieve.}
Having formalized memorization and retrieval via our regression framework, we can now combine the two to produce sequence layers that \emph{directly performs associative recall as a forward pass}.
Recall that we are given $\{\qbf_i\}_{i=1}^T$, $\{\kbf_i\}_{i=1}^T$, and $\{\vbf_i\}_{i=1}^T$, and we seek to produce a set of outputs $\{\ybf_i\}_{i=1}^T$ with our sequence layer.
The forward pass of a test-time regression layer first \emph{memorizes} a set of key-value pairs and then produces $\ybf_t$ by \emph{retrieving} a value from its associative memory.
More specifically, in causal modeling, at each timestep $t$, we first solve \autoref{eqn:general-least-squares} to get a regressor $\memory_t$ that memorizes the prefix $\{(\kbf_i, \vbf_i)\}_{i=1}^t$; then we apply $\memory_t$ to $\qbf_t$ to get $\ybf_t = \memory_t(\qbf_t)$.
This naturally generalizes to non-causal tasks where $\memory_t$ can depend on suffix key-value pairs $\{\kbf_i,\vbf_i\}_{i > t}$ as well.
Our procedure is compatible with multihead design patterns inspired by multihead attention \citep{Vaswani2017AttentionAllYou}, multi-query attention \citep{Shazeer2019FastTransformerDecoding}, and grouped query attention \citep{Ainslie2023GQATrainingGeneralized}.
For example, multihead test-time regression can be implemented using $H$ independent sets of queries, keys, and values, resulting in $H$ independent associative memory maps at each timestep.

\paragraph{A recipe for designing your own sequence layer.}
Although \autoref{eqn:general-least-squares} succinctly summarizes the memorization step of our framework, it leaves many details unspecified.
For example, over which class of regressors $\Mcal$ should we consider?
Additionally, which associations should we prioritize memorizing?
Finally, which optimization procedure should we use to minimize the loss?
The regression-memory correspondence thus allows for a vast design space.
We partition this wide landscape into three simple design choices:
\begin{enumerate}
    \item the relative importance of each association, specified through weights in  \autoref{eqn:general-least-squares},
    \item the function class $\Mcal$ over which we search,
    \item the minimization algorithm,
\end{enumerate}
giving us a formulaic ``recipe'' for deriving sequence layers that can perform associative recall.

\section{Deriving existing architectures from regression}\label{sec:deriving-architectures}

To demonstrate how our framework unifies our understanding of sequence models, we succinctly derive a swath of recently proposed sequence layers.
We categorize existing sequence layers into four distinct classes: linear attention (possibly with feature maps), linear attention with gating (which includes state-space models like Mamba), fast-weight layers, online learners, and softmax self-attention.

\paragraph{Notation.}
We first explain the mathematical notation we will use in our derivations.
We use bold lowercase letters to refer to vectors (e.g. $\xbf, \kbf, \vbf$) and bold uppercase letters to refer to matrices (e.g. $\Mbf, \Kbf, \Vbf$).
Assume we have a sequence of $T$ inputs $\xbf_1, \ldots, \xbf_T$ which have already transformed into a sequence of keys and values $(\kbf_1, \vbf_1), \ldots, (\kbf_T, \vbf_T)$.
We will use these keys and values as the dataset for our test-time regression problem.
Let $\Kbf_t = \bmat{\kbf_1 \ldots \kbf_t}\transpose \in \reals^{t\times \dkey}$ and $\Vbf_t = \bmat{\vbf_1 \ldots \vbf_t}\transpose \in \reals^{t\times \dval}$ be matrices that contain $t$ keys and $t$ values respectively, for $t=1, \ldots, T$.
Each key/value vector is a row of the key/value matrix.

\subsection*{Vignette 1: Linear attention is (suboptimal) linear least squares regression}
We start by deriving an architecture from linear least squares, the most classic regression method.
Linear least squares corresponds to the following choices in our recipe.
\begin{enumerate}
    \item Choice of weights: assign equal weight $\gamma_i^{(t)} = 1$ to each association,  \begin{align}\memory_{\textsc{linear},\,t} = \argmin_{\memory\in \linearfunctions} \frac{1}{2}\sum_{i=1}^t \norm{\vbf_i - \memory(\kbf_i)}_2^2.\label{eqn:linear-least-squares}\end{align}
    \item Choice of function class: linear functions $\linearfunctions = \{\memory\mid \memory(\kbf) = \Mbf \kbf,\ \Mbf \in \reals^{\dval\times\dkey}\}$.
    \item Choice of minimization algorithm: analytical solution equivalent to one step of Newton's method\footnote{Since the objective function is quadratic, the analytical solution is equivalent to one iteration of Newton's method initializing from the origin.}.
\end{enumerate}
Since $\memory_{\textsc{linear},\,t}$ is a linear function, it is equivalent to a linear map $\Mbf_t\in \reals^{\dval\times\dkey}$.
Minimizing the linear least squares objective results in the solution:
\begin{align}
\Mbf_t = \argmin_{\Mbf} \frac{1}{2}\sum_{i=1}^t \norm{\vbf_i - \Mbf\kbf_i}_2^2
    =\begin{cases}
        \Vbf_t\transpose \Kbf_t(\Kbf_t\transpose\Kbf_t)\inverse, \ \text{overdetermined, e.g. } t \geq \dkey, \label{eqn:ols}\\
        \Vbf_t\transpose (\Kbf_t\Kbf_t\transpose)\inverse\Kbf_t, \ \text{underdetermined, e.g. } t < \dkey, %
    \end{cases}
\end{align}
where we choose the min-norm solution in the underdetermined case.
The overdetermined case corresponds to $\Kbf_t$ having rank $\dkey$ (linearly independent columns) and the underdetermined case corresponds to $\Kbf_t$ having rank $t$ (linearly independent rows).
In general, we can express the solution in terms of a Moore-Penrose pseudoinverse: $\Mbf_t = \Vbf_t\transpose(\Kbf_t\pinv)\transpose$.
We restrict our discussion to the case of $t \geq \dkey$ with linearly independent columns, which is the case for typical applications.

\paragraph{Relationship to linear attention.}
Although $\memory_{\textsc{linear},\,t}$ is the optimal linear memory with respect to our objective function, it can be expensive to compute, requiring us to invert a $\dkey\times\dkey$ matrix at every timestep.
One crude but hardware-efficient simplification would be to assume that $\Kbf_t\transpose\Kbf_t = \Ibf$, avoiding matrix inversion completely, at the cost of some recall ability.
After making this rough approximation, \emph{we arrive exactly at the equations for linear attention}:
\begin{align}
\ybf_t = \memory_{\textsc{linear}}(\qbf_t)  = \Vbf_t\transpose \Kbf_t(\Kbf_t\transpose\Kbf_t)\inverse \qbf_t \approx \Vbf_t\transpose\Kbf_t\qbf_t =\sum_{i=1}^t \vbf_i\kbf_i\transpose \qbf_t.
\end{align}

Our derivation clarifies exactly how linear attention underperforms as an associative memory.
When the keys are zero mean, $\Kbf_t\transpose\Kbf_t$ is proportional to the empirical covariance matrix of the keys.
Thus linear attention \emph{is a crude associative memory that ignores the covariance between the dimensions of the key vectors}.
In fact, it is an optimal linear memory only when $t \leq \dkey$ and the keys are orthonormal $\kbf_i\transpose\kbf_j = \delta_{ij}$.

From a training stability perspective, the inverse covariance factor helps prevent the output $\ybf_t$ from becoming unbounded.
We can bound the norm of the output of a least squares layer (\autoref{eqn:ols}) by
\begin{align}
   \norm{\ybf_t}_2 = \norm{\Vbf_t\transpose\Kbf_t(\Kbf_t\transpose\Kbf_t)\inverse \qbf_t}_2 \leq \frac{\norm{\qbf_t}_2\left(\sum_{i=1}^t\norm{ \vbf_i}_2\norm{\kbf_i}_2\right)}{\lambda_{\text{min}}(\Kbf_t\transpose\Kbf_t)}
\end{align}
where $\lambda_{\text{min}}(\Kbf_t\transpose\Kbf_t)$ is the minimum eigenvalue of $\Kbf_t\transpose\Kbf_t$.
See \autoref{sec:linear-regression-bound} for the derivation.
When we ignore the covariance matrix, approximating it with the identity, the denominator disappears and the output $\ybf_t$ can grow arbitrarily large with the sequence length, leading to training instability.
This instability indeed shows up in practice for linear attention, and \citet{Qin2022DevilLinearTransformer} proposed output normalization to stabilize training.
Our derivation shows that output normalization works by approximately restoring the self-normalizing property of \autoref{eqn:ols} that linear attention loses.

Regardless of training stability, we will see in \autoref{sec:key-construction} that ignoring the covariance causes linear attention to perform worse at associative recall.

\paragraph{Efficient computation.}
Conveniently, we can compute $\Mbf_t$ in \autoref{eqn:ols} for $t=1, \ldots T$ in parallel or sequentially.
To compute them in parallel, we solve a batch of $T$ independent linear systems.
To compute them sequentially, we note that $(\Kbf_t\transpose\Kbf_t)\inverse = (\Kbf_{t-1}\transpose \Kbf_t + \kbf_t\kbf_t\transpose)\inverse$, allowing us to autoregressively compute the inverses via the Woodbury matrix identity.
The sequential form is known as online linear regression or recursive least squares (RLS) \citep{Haykin2014AdaptiveFilterTheory}. %
Thus during training we can parallelize across the sequence axis by solving a batch of $T$ linear systems, and during testing we can unroll the recurrence into a recurrent neural network.
In prior deep learning literature, this idea of linear-regression-as-a-layer has been explored by \citet{Garnelo2023ExploringSpaceKeyValueQuery} and \citet{vonOswald2023UncoveringMesaoptimizationAlgorithms} under the names ``intention'' and ``mesa-layer'', respectively, motivated by heuristic similarities to softmax attention, rather than deriving them from first principles via enforcing associative recall as we do here. %

\paragraph{Nonlinear associative memory.}
So far we have only considered \emph{linear} associative memory implemented by linear functions in $\linearfunctions$, which have limited expressiveness.
We can derive \emph{nonlinear} memory by using a feature map $\phi: \reals^\dkey \to \reals^\dphi$, replacing each key $\kbf_i$ with $\phi(\kbf_i)$.
Then our function class is defined by $\Mcal_{\phi} = \{\memory \mid \memory(\kbf) = \Mbf\phi(\kbf),\ \Mbf \in \reals^{\dval\times\dphi}\}$ and we solve the least-squares objective $\argmin_{\Mbf} \sum_{i=1}^t\norm{\vbf_i - \Mbf \phi(\kbf_i)}_2^2/2$.
Solving this least squares objective for $\Mbf$ results in an optimal associative memory that is nonlinear with respect to the input keys.
Parameterizing associative memory with a nonlinear feature map has a long history, dating back to \citet{Poggio1975OptimalNonlinearAssociative} which found polynomial feature maps to be highly effective, consistent with more recent works \citep{Zhang2023HedgehogPorcupineExpressive,Arora2024SimpleLinearAttention}.

If we make the same crude approximation as linear attention of disregarding the covariance among $\{\phi(\kbf_i)\}_{i=1}^t$, we arrive at a suboptimal nonlinear memory.
Thus, recent works that apply linear attention with a nonlinear feature map, such as \citet{Schlag2021LinearTransformersAre,Kasai2021FinetuningPretrainedTransformers,Zhang2023HedgehogPorcupineExpressive,Chen2024DiJiangEfficientLarge}, and others, can be understood within our framework as suboptimal nonlinear memory as well.

\subsection*{Vignette 2: Gated linear attention and state-space models are (suboptimal) weighted linear least squares regression}

For many kinds of sequential data, such as time-series data and text data, recent tokens are more informative of predicting the next token.
Thus a natural requirement is to encourage our memory $\memory_t$ to focus more on the recent history by downweighting the past.
Fortunately, this can be straightforwardly enforced through classic \emph{weighted} linear least squares regression together with geometrically decaying weights to \emph{forget} older key-value pairs.

To develop an architecture from weighted linear least squares, we only need to add weights to the optimization objective relative to ordinary least squares:
\begin{enumerate}
    \item Choice of weights: assign time-decaying weights to associations,  \begin{align}\memory_{\textsc{linear},\,t} = \argmin_{\memory\in \linearfunctions} \frac{1}{2}\sum_{i=1}^t \gamma_{i}^{(t)}\norm{\vbf_i - \memory(\kbf_i)}_2^2,\label{eqn:decaying-ols}\end{align}
    Here, to connect to past literature, we focus on weights are geometrically decaying in time: $\gamma_{i}^{(t)} = \prod_{j=i+1}^t \gamma_j$ and $\gamma_i \in [0, 1]$ for $i=1, \ldots, t$. Each association is down-weighted based on how far it is from the current timestep $t$: $\gamma_1^{(t)} \leq \ldots \gamma_{t-1}^{(t)} \leq \gamma_t^{(t)} = 1$.
    \item Choice of function class: linear functions $\linearfunctions = \{\memory\mid \memory(\kbf) = \Mbf \kbf,\ \Mbf \in \reals^{\dval\times\dkey}\}$.
    \item Choice of minimization algorithm: analytical solution equivalent to one step of Newton's method.
\end{enumerate}
We can solve weighted linear regression by instead solving an unweighted regression problem involving rescaled keys and values, $\Vbf_t\mapsto \Gammabf_t^{1/2}\Vbf_t$ and $\Kbf_t\mapsto \Gammabf_t^{1/2}\Kbf_t$, where we defined a diagonal matrix $\Gammabf_t\in \reals^{t\times t}$ with entries $(\Gammabf_t)_{ii} = \gamma_{i}^{(t)}$.
Making the rescaled substitution and reusing the analytical min-norm solution that we had before, we see that \begin{align}
    \Mbf_t &= 
    \begin{cases}
        \Vbf_t\transpose \Gammabf_t \Kbf_t(\Kbf_t\transpose\Gammabf_t\Kbf_t)\inverse, \ \text{overdetermined, e.g. }\, t \geq \dkey\\
        \Vbf_t\transpose (\Kbf_t\Kbf_t\transpose)\inverse\Kbf_t, \ \text{underdetermined, e.g. }\ t < \dkey.
    \end{cases}\label{eqn:optimal-linear-memory-weighted}
\end{align}
The underdetermined case has the same solution in both the weighted and unweighted case; intuitively this is because when the system is underdetermined, there are infinitely many solutions that interpolate the data points and attain zero regression loss, rendering the weights irrelevant.

\paragraph{Relationship to linear attention with gating and state-space models.}
Once again, computing $\Mbf_t$ requires matrix inversion which is not efficient on matrix-multiplication based hardware.
Making a crude approximation similar to before of $\Kbf_t\transpose\Gammabf_t\Kbf_t\approx \Ibf$ results in $\Mbf_t \approx \Vbf_t\transpose \Gammabf_t\Kbf_t = \sum_{i=1}^t \gamma_i^{(t)} \vbf_i\kbf_i\transpose$.
We can further unroll this equation into a recurrence due to the multiplicative structure of our weights:
\begin{align}
    \Mbf_t \approx \sum_{i=1}^t \gamma_i^{(t)} \vbf_i\kbf_i\transpose
    = \sum_{i=1}^t \left(\prod_{j=i+1}^t \gamma_j\right) \vbf_i\kbf_i\transpose 
    = \gamma_t \sum_{i=1}^{t-1} \gamma_i^{(t-1)} \vbf_i\kbf_i\transpose  + \vbf_t\kbf_t\transpose
    = \gamma_t \Mbf_{t-1} + \vbf_t\kbf_t\transpose.
\end{align}
This is exactly the recurrence implemented by gated variants of linear attention, such as GLA \citep{Yang2024GatedLinearAttention}, RetNet \citep{Sun2023RetentiveNetworkSuccessor}, Gateloop \citep{Katsch2024GateLoopFullyDataControlled}, RWKV-6 \citep{Peng2024EagleFinchRWKV}, mLSTM \citep{Beck2024XLSTMExtendedLong}, LRU \citep{Orvieto2023ResurrectingRecurrentNeural}, as well as state-space models like Mamba-2 \citep{Dao2024TransformersAreSSMs}.
Forgetting older tokens significantly improves performance, which was first discovered \emph{empirically} \citep{Hochreiter1997LongShortTermMemory,Greff2017LSTMSearchSpace,vanderWesthuizen2018UnreasonableEffectivenessForget}.
Our derivation of gating from weighted least squares provides a mathematical grounding for the forgetting heuristic upon which these sequence layers were based.

\paragraph{Efficient computation.}
Like before, we can compute the memory of our architecture (without approximation) sequentially or in parallel as long as we are willing to invest FLOPs in matrix inversion.
Parallel computation can be done by solving a batch of $T$ $\dkey\times \dkey$ linear systems.
The sequential version can be derived by applying the Woodbury matrix inversion formula while taking into account our geometrically-decaying weights.
Geometrically-weighted recursive least squares (RLS) is a classic method from the adaptive filtering literature \citep{Johnstone1982ExponentialConvergenceRecursive}.

\subsection*{Vignette 3: Fast weight programmers and online learners are first-order methods for solving streaming least-squares}
Up until now we have derived linear associative memory architectures through direct analytical solution, equivalent to one step of Newton's method, a second-order optimization method.
However, second-order iterative methods are computationally expensive, requiring matrix inversions at each step.
On the other hand, disregarding the matrix inversion as done by linear attention leads to poor performance, since it ignores the covariance structure of the data.

A computationally cheaper alternative is to apply a first-order gradient-descent method. We consider this here with the following design choices:
\begin{enumerate}
    \item Choice of weights: assign equal weight $\gamma_i^{(t)} = 1$ to each association,  \begin{align}\memory_{\textsc{linear},\,t} = \argmin_{\memory\in \linearfunctions} \frac{1}{2}\sum_{i=1}^t \norm{\vbf_i - \memory(\kbf_i)}_2^2.\label{eqn:least-squares-batch-but-online}\end{align}
    \item Choice of function class: linear functions $\linearfunctions = \{\memory\mid \memory(\kbf) = \Mbf \kbf,\ \Mbf \in \reals^{\dval\times\dkey}\}$.
    \item Choice of minimization algorithm: gradient descent, possibly with adaptive step sizes, momentum, or weight decay (L2 regularization).
\end{enumerate}
We can now derive the test-time regression layer corresponding to a first-order iterative method.

Since we consider linear functions, solving \autoref{eqn:least-squares-batch-but-online} is equivalent to minimizing the objective
\begin{align}
\Lcal_t(\Mbf) = \sum_{i=1}^t L_i(\Mbf) = \sum_{i=1}^t \frac{1}{2}\norm{\vbf_i - \Mbf\kbf_i}_2^2, \label{eqn:unregularized-linear-least-squares}
\end{align}
using its gradient
\begin{align}
    \grad\Lcal_t(\Mbf) &= \sum_{i=1}^t \grad L_i(\Mbf) = \sum_{i=1}^t (\Mbf\kbf_i - \vbf_i)\kbf_i\transpose = (\Mbf\Kbf_t\transpose - \Vbf_t\transpose)\Kbf_t.
\end{align}
to find $\Mbf_t$.
Let the initialization be at the origin $\Mbf^{(0)}_t=\zerovec$.
Iteratively solving this objective with full-batch gradient descent and step size $\beta_t^{(i)}$ at each iteration results in
\begin{align}
    \Mbf^{(i)}_t &= \Mbf^{(i-1)}_t - \beta_t^{(i)}\grad\Lcal_t(\Mbf^{(i-1)}_t) \\
           &= \Mbf^{(i-1)}_t - \beta_t^{(i)}\sum_{j=1}^t \grad L_j(\Mbf^{(i-1)}_t) \label{eqn:batch-gradient-descent}\\
           &= \Mbf^{(i-1)}_t - \beta_t^{(i)}\sum_{j=1}^t (\Mbf^{(i-1)}_t\kbf_j - \vbf_j)\kbf_j\transpose \\
           &= \Mbf^{(i-1)}_t(\Ibf - \sum_{j=1}^t\beta_t^{(i)} \kbf_j\kbf_j\transpose) + \beta_t^{(i)} \sum_{j=1}^t \vbf_j\kbf_j\transpose.
\end{align}
After $I$ iterations, we produce the linear memory represented by $\Mbf_t = \Mbf^{(I)}_t$.

\paragraph{Linear attention and variants as one-step of batch gradient descent}
Consider the $I=1$ case with step size $\beta_t^{(i)} = 1$.
Since we initialize at the origin, we obtain the equation for linear attention 
\begin{align}
    \Mbf^{(1)}_t = \Mbf^{(0)}_t - \grad\Lcal_t(\Mbf^{(0)}_t) = \Vbf_t\transpose\Kbf_t = \sum_{i=1}^t \vbf_i \kbf_i\transpose. \label{eqn:linear-attention-gd}
\end{align}
This is exactly the result of performing \emph{one-step of full batch gradient descent} on the least square objective $\Lcal_t$. 
A similar result can be shown for layers that implement linear attention with forgetting, performing one step of full batch gradient descent on the weighted least square objective.

A standard technique to accelerate the convergence of gradient descent is to precondition the gradient \citep{Boyd2004ConvexOptimization}.
The optimal preconditioner in this case, due to the quadratic objective, is the \emph{constant} Hessian of $\Lcal_t$, $\Hbf_{\Lcal_t}(\Mbf)=\Kbf_t\transpose\Kbf_t$.
This preconditioner allows gradient descent to converge in a \emph{single update}
\begin{align}
    \Mbf^{(1)}_t = \Mbf^{(0)}_t - \grad \Lcal_t(\Mbf^{(0)}_t)\Hbf_{\Lcal_t}(\Mbf^{(0)}_t)\inverse = \Vbf_t\transpose\Kbf_t(\Kbf_t\transpose\Kbf_t)\inverse,
\end{align}
reproducing the analytic solution from \autoref{eqn:ols}.
The preconditioned recurrence is equivalent to one step of Newton's method, a second order iterative method.

By taking this iterative optimization perspective, we see that \emph{linear attention and the ideal linear associative memory lie on two extremes: one-step of batch gradient descent with no preconditioning (\autoref{eqn:linear-attention-recurrence}) versus one-step of batch gradient descent with perfect preconditioning (\autoref{eqn:ols})}.
The difference in performance comes from whether the optimizer accounts for the curvature of the objective, governed by the covariance of the keys \citep{Boyd2004ConvexOptimization}.
Exploring other preconditioners, e.g. quasi-Newton methods, that interpolates between the two extremes may be an interesting direction for future research.

\paragraph{Stochastic gradient descent as a layer.}
An alternative to full-batch gradient descent is stochastic gradient descent (SGD) where we use a subset of the samples to compute the gradient at each step.
This is particularly convenient in causal sequence modeling where keys and values typically are presented one at a time, rather than all at once.
In this streaming setting, we can instead perform single-example SGD to iteratively minimize \autoref{eqn:general-least-squares}.
In the case of linear memories $\linearfunctions$, \autoref{eqn:batch-gradient-descent} with a single-example and variable step sizes $\beta_t^{(i)}$ becomes
\begin{align}
    \Mbf_t^{(i)} &= \Mbf_t^{(i-1)} - \beta_t^{(i)} \grad L_t(\Mbf_{t}^{(i-1)}) \label{eqn:sgd} \\
    &=\Mbf_{t}^{(i-1)} + \beta_t^{(i)} (\vbf_t - \Mbf_{t}^{(i-1)}\kbf_t)\kbf_t\transpose \label{eqn:sgd-2} \\
    &= \Mbf_{t}^{(i-1)}(\Ibf - \beta_t^{(i)}\kbf_t\kbf_t\transpose) + \beta_t^{(i)} \vbf_t\kbf_t\transpose, 
\end{align}
where $\Mbf_t$ is the matrix representation of $\memory_t$.
When we take only one gradient update per timestep $t$ and initialize our iterate using the previous timestep's linear map, $\Mbf_t^{(0)} = \Mbf_{t-1}$, we end up with the recurrence of DeltaNet \citep{Schlag2021LinearTransformersAre,Yang2024ParallelizingLinearTransformers} from \autoref{eqn:delta-rule}:
\begin{align}
    \Mbf_t = \Mbf_t^{(1)} = \Mbf_{t-1}(\Ibf - \beta_t\kbf_t\kbf_t\transpose) + \beta_t \vbf_t\kbf_t\transpose\label{eqn:linear-sgd}.
\end{align}
As we perform stochastic gradient descent, we produce an online sequence $\{\Mbf_i\}_{i=1}^t$, each of which is an associative memory $\memory_i$ for key-value pairs up to $(\kbf_i, \vbf_i)$.
The stochastic gradient approach to streaming inputs has the advantage that $\memory_t$ focuses more on recent key-value pairs, rather than uniformly over all timesteps as in the case of unweighted RLS and unweighted linear attention.

\autoref{eqn:linear-sgd} is known by many names, such as the delta rule or the Widrow-Hoff algorithm \citep{Widrow1988AdaptiveSwitchingCircuits}.
Within the adaptive filtering literature this recurrence is known as Least-Mean Squares (LMS) \citep{Haykin2014AdaptiveFilterTheory}, serving as one of the most common algorithms for streaming signals processing.
From this derivation, we see that the prevalence of the update across disciplines comes from the fact that the update rule is actually an SGD update.

In our case of sequence layers, the SGD update occurs \emph{within} the layer as part of the forward pass.
One extension to \autoref{eqn:linear-sgd} is to consider SGD with nonlinear parametric regressors $\memory_t$.
Indeed, this extension is equivalent to test-time training (TTT) layers proposed by \citet{Sun2024LearningLearnTest}, which apply SGD to a nonlinear memory $\memory_t$, e.g. multilayer perceptrons or linear maps with layer normalizations.
In contrast to our associative memory derivation here, \citet{Sun2024LearningLearnTest} motivated their method by viewing the squared loss as an self-supervised auto-encoding objective.
Another extension is to take multiple gradient steps per example $(\kbf_t, \vbf_t)$, which was recently explored by DeltaProduct \citep{Siems2025DeltaProductIncreasingExpressivity}.

\paragraph{Including adaptive step sizes.}
To further improve the performance of a gradient-descent-based sequence layer, we can consider adaptive step sizes.
A common way to derive adaptive step sizes is to use the relationship between gradient descent and online learning \citep{Cesa-Bianchi2004GeneralizationAbilityOnline,Zinkevich2003OnlineConvexProgramming}, a connection that has produced many successful algorithms in the past, such as Adagrad \citep{Duchi2011AdaptiveSubgradientMethods} and Adam \citep{Kingma2015AdamMethodStochastic}.

Consider the Longhorn sequence layer derived by \citet{Liu2024LonghornStateSpace} by solving the online learning objective
\begin{align}
\Mbf_{t} = \argmin_\Mbf\ \frac{1}{2}\norm{\Mbf - \Mbf_{t-1}}_F^2 + \frac{\delta_t}{2} \norm{\vbf_t - \Mbf \kbf_t}_2^2.\label{eqn:longhorn-objective}
\end{align}
This corresponds to $d(\Mbf_{t-1}, \Mbf) = \norm{\Mbf - \Mbf_{t-1}}_F^2/2$ and $\ell_t(\Mbf) = \delta \norm{\vbf_t - \Mbf \kbf_t}_2^2/$ in \autoref{eqn:online-learning-objective}.
The solution to \autoref{eqn:longhorn-objective} is the recurrence
\begin{align}
    \Mbf_{t} &= \Mbf_{t-1}\left(\Ibf - \frac{\delta_t}{1 + \delta_t\norm{\kbf_t}_2^2} {\kbf_t\kbf_t\transpose}\right) + \frac{\delta_t}{1 + \delta_t \norm{\kbf_t}_2^2}\vbf_t\kbf_t\transpose.\label{eqn:longhorn-update}
\end{align}
However, by comparing the Longhorn update in \autoref{eqn:longhorn-update} to the SGD update in \autoref{eqn:linear-sgd}, we see that Longhorn's recurrence implements SGD but with a conservative adaptive step size of $\beta_t = \delta_t / (1 + \delta_t\norm{\kbf_t}_2^2)$.
Thus although the Longhorn layer was derived from an online learning perspective, its global objective is still the least squares regression objective.

We can better understand the effects of this adaptive step size by looking at $\delta_t$, which controls how much we care about memorizing the latest association $(\kbf_t, \vbf_t)$.
When $\delta_t = 0$, we see that $\Mbf_t = \Mbf_{t-1}$ and the latest association is excluded completely.
When $\delta_t \to \infty$, \autoref{eqn:longhorn-objective} becomes equivalent to a constrained optimization problem
\begin{align}
\Mbf_{t} = \argmin_\Mbf\ \frac{1}{2}\norm{\Mbf - \Mbf_{t-1}}_F^2\quad \subjectto \Mbf_t\kbf_t = \vbf_t.\label{eqn:nlms-objective}
\end{align}
Solving this constrained optimization problem results in a hyperparameter-free step size $\beta_t=1/\norm{\kbf_t}_2^2$ with update
\begin{align}
    \Mbf_{t} &= \Mbf_{t-1}\left(\Ibf - \frac{1}{\norm{\kbf_t}_2^2} {\kbf_t\kbf_t\transpose}\right) + \frac{1}{ \norm{\kbf_t}_2^2}\vbf_t\kbf_t\transpose.\label{eqn:nlms-update}
\end{align}
This special case is also known as the normalized LMS (NLMS) update in adaptive filtering \citep{Castoldi2009MinimumdisturbanceDescriptionDevelopment} and known as the projection method \citep{Tanabe1971ProjectionMethodSolving} or Kaczmarz's method \citep{Kaczmarz1937AngenaherteAuflosungSystemen} in optimization theory.
The advantage of \autoref{eqn:nlms-update} is that it is hyperparameter-free, which may make it easier to tune.
However, to our knowledge, the recurrence in \autoref{eqn:nlms-update} has not been explored as a sequence layer.

\paragraph{Including L2 regularization.}
In addition to adaptive step sizes, we can add regularization.
Consider adding L2 regularizing to the least squares objective \autoref{eqn:unregularized-linear-least-squares}
\begin{align}
    \Lcal_{\text{reg}, t}(\Mbf) = \sum_{i=1}^t L_i(\Mbf) + \frac{\lambda_i}{2}\norm{\Mbf}_2^2
\end{align}
where $\lambda_i\in [0, 1]$ is the regularization strength.
When we minimize this regularized objective with single-example stochastic gradient descent like in \autoref{eqn:linear-sgd}, we obtain the recurrence
\begin{align}
    \Mbf_t &= \Mbf_{t-1} -\beta_t \left[\grad L_t(\Mbf_{t-1}) + \lambda_t \Mbf_{t-1}\right] \\
    &= (1 - \beta_t\lambda_t)\Mbf_{t-1} -\beta_t \grad L_t(\Mbf_{t-1}) \\
    &= (1 - \beta_t\lambda_t) \Mbf_{t-1} + \beta_t(\vbf_t - \Mbf_{t-1}\kbf_t)\kbf_t\transpose,\label{eqn:sgd-weight-decay-recurrence}
\end{align}
which is known as Leaky LMS in adaptive filtering \citep{Haykin2014AdaptiveFilterTheory}.
Notably, the first term of our new recurrence resembles \autoref{eqn:sgd-2} but also includes a multiplicative factor similar to a forget gate.

Recently, \citet{Yang2024GatedDeltaNetworks} proposed Gated DeltaNet motivated by combining DeltaNet \citep{Schlag2021LinearTransformersAre} with the forget gating of Mamba-2 \citep{Dao2024TransformersAreSSMs}, following the online learning perspective of \citet{Liu2024LonghornStateSpace}.
Here we show that the Gated DeltaNet recurrence is actually an equivalent reparameterization of our recurrence in \autoref{eqn:sgd-weight-decay-recurrence}. Starting with \autoref{eqn:sgd-weight-decay-recurrence}, we have
\begin{align}
    \Mbf_t &= (1 - \beta_t\lambda_t) \Mbf_{t-1} + \beta_t(\vbf_t - \Mbf_{t-1}\kbf_t)\kbf_t\transpose \\
    &= \Mbf_{t-1}\left[ \underbrace{( 1 - \beta_t \lambda_t)}_{\alpha_t}\Ibf - \beta_t\kbf_t\kbf_t\transpose \right] + \beta_t \vbf_t\kbf_t\transpose \\
    &= \Mbf_{t-1}\alpha_t\left[\Ibf - \underbrace{\frac{\beta_t}{\alpha_t}}_{\eta_t}\kbf_t\kbf_t\transpose \right] + \frac{\beta_t}{\alpha_t} \underbrace{(\alpha_t \vbf_t)}_{\vbf'_t}\kbf_t\transpose \\
    &= \Mbf_{t-1}\alpha_t\left(\Ibf - \eta_t\kbf_t\kbf_t\transpose \right) + \eta_t \vbf'_t\kbf_t\transpose \quad \text{(Gated DeltaNet recurrence)}\label{eqn:gated-deltanet-recurrence}
\end{align}
Thus Gated DeltaNet is equivalent to L2-regularized regression via SGD where the regularization strength $\lambda_t$ is coupled with the step size $\beta_t$ and $\vbf_t$ is rescaled by $\alpha_t$.
We can transform \autoref{eqn:gated-deltanet-recurrence} into \autoref{eqn:sgd-weight-decay-recurrence} via the invertible mapping $(\alpha_t, \eta_t, \vbf_t') = (1 - \beta_t\lambda_t, \beta_t/(1 - \beta_t\lambda_t), (1 - \beta_t\lambda_t)\vbf_t)$.
The inverse map is given by $(\lambda_t, \beta_t, \vbf_t) = ((1-\alpha_t)/(\alpha_t\eta_t), \alpha_t\eta_t,\allowbreak \vbf'_t/\alpha_t)$.
Past works have shown that decoupling weight decay from the step size results in better generalization and easier-to-tune hyperparameters \citep{Loshchilov2019DecoupledWeightDecay}, suggesting that similar modifications to Gated DeltaNet may improve model performance.

\subsection*{Vignette 4: Softmax attention is nonparametric regression}\label{vig:softmax-attention-is-nonparametric}
Having exhaustively derived multiple associative memory architectures via linear and nonlinear parametric regression, we can now derive self-attention-like architectures using the tools of nonparametric regression.

\paragraph{Unnormalized softmax attention is (suboptimal) kernel regression.}
We start first from kernel regression as the most common nonparametric regression method \citep{Scholkopf2001LearningKernelsSupport}, and applying the kernel trick to our feature-mapped linear memory.
Let $k:\reals^{\dkey\times \dkey} \to \reals$ be a Mercer kernel such that $k(\kbf_i, \kbf_j) = \phi(\kbf_i)\transpose\phi(\kbf_j) = (\Phi\Phi\transpose)_{ij}$ where $\Phi = \bmat{\phi(\kbf_1) \ldots \phi(\kbf_t)}\transpose \in \reals^{t\times \dphi}$ and $\phi: \reals^\dkey\to \reals^\dphi$ is a possibly  infinite-dimensional feature map.
Following our recipe for model design, we make the following design choices:
\begin{enumerate}
    \item Choice of weights: assign equal weight $\gamma_i^{(t)} = 1$ to each association  \begin{align}\memory_{k,\,t} = \argmin_{\memory\in \Mcal_k} \frac{1}{2}\sum_{i=1}^t \norm{\vbf_i - \Mbf \phi(\kbf_i)}_2^2,\end{align} where $\phi$ is possibly an infinite-dimensional feature map.
    \item Choice of function class: linear functions $\Mcal_{k}= \{\memory\mid \memory(\kbf) = \Mbf \phi(\kbf),\ \Mbf \in \reals^{\dval\times\dphi}\}$.
    \item Choice of minimization algorithm: analytical solution.
\end{enumerate}

The output of the optimal kernelized nonlinear associative memory induced by the kernel $k$ at timestep $t$ is the standard kernel regression prediction equation \citep{Scholkopf2001LearningKernelsSupport}
\begin{align}
   \memory_{k,\, t}(\qbf) = \Vbf_t\transpose k(\Kbf_t, \Kbf_t)\inverse k(\Kbf_t, \qbf).
\end{align}
Here we follow standard shorthand notation from the kernel learning literature: $k(\Kbf_t, \Kbf_t)$ is a shorthand for the matrix with $k(\kbf_i, \kbf_j)$ in entry $(i,j)$; $k(\Kbf_t, \qbf)$ is a shorthand for the vector with $k(\kbf_i, \qbf)$ in entry $i$.
When we use an exponential kernel $k(\kbf_i, \kbf_j) = \exp(\kbf_i\transpose\kbf_j/\sqrt{\dkey})$, corresponding to an infinite-dimensional feature map, and make the crude approximation $k(\Kbf_t, \Kbf_t) \approx \Ibf$, we obtain
\begin{align}
    \memory_{k, t}(\qbf) &= \Vbf_t\transpose k(\Kbf_t, \Kbf_t)\inverse k(\Kbf_t, \qbf) \\
    &\approx \Vbf_t\transpose k(\Kbf_t,\qbf) \\
    &= \sum_{i=1}^t \vbf_i\, k(\kbf_i, \qbf) \\
    &= \sum_{i=1}^t \vbf_i \exp\left(\frac{\kbf_i\transpose\qbf}{\sqrt{\dkey}}\right),
\end{align}
an associative memory map equivalent to an \emph{unnormalized} softmax attention.
This relationship with kernel regression has been used by many previous works to derive more efficient versions of self-attention \citep{Peng2020RandomFeatureAttention,Chen2021SkyformerRemodelSelfattention,Choromanski2020RethinkingAttentionPerformers,Lu2021SOFTSoftmaxfreeTransformer,Chen2024DiJiangEfficientLarge}, borrowing techniques from kernel approximation methods.
However unnormalized softmax attention is empirically known to be unstable to train \citep{Lu2021SOFTSoftmaxfreeTransformer}.
Thus we do not stop our derivations here; instead, we will see how the standard normalized softmax attention comes from another kind of nonparametric regressor.

\paragraph{Softmax attention with QKNorm is a locally-constant nonparametric regressor.}
Local polynomial regression is another classic class of nonparametric estimators \citep{Wasserman2006AllNonparametricStatistics}, which have been historically used as nonlinear smoothing functions in data analysis \citep{Fan2018LocalPolynomialModelling}.
Here, we consider the function class $\Mcal^p$, which is the set of all vector-valued functions that are \emph{locally} approximated by a degree $p$ polynomial around each point $\qbf \in \reals^\dkey$. Here locality is measured by some similarity function $s$ that only depends on the distance to $\qbf$, i.e. $s(\kbf, \qbf) = s(\norm{\kbf - \qbf}_2)$.
A classic example is the exponential smoothing kernel $s_{\text{exp},\, B}(\kbf, \qbf) = \exp(-\norm{\kbf - \qbf}_2^2/B)$ where $B$ is a parameter controlling the local neighborhood size around $\qbf$.
Local polynomial regression is most often discussed for one-dimensional data, but here we generalize it to the multivariate case.

Nonparametric regression with local polynomial estimators corresponds to the following design choices:
\begin{enumerate}
    \item Choice of weights: assign weights to associations based on similarity to $\qbf$, 
    \begin{align} \memory_{p,\,t}(\qbf) = \argmin_{\memory\in \Mcal^p} \frac{1}{2}\sum_{i=1}^t \gamma_i^{(t)} \norm{\vbf_i - \memory(\kbf_i)}_2^2 \label{eqn:local-polynomial-loss}\end{align}
    with weights given by $\gamma_i^{(t)}=s(\kbf_i, \qbf)=s(\norm{\kbf_i - \qbf}_2)$.%
    \item Choice of function class:  $\Mcal^p$, the class of functions $\memory_p$ such that for all points $\qbf\in \reals^\dkey$, the function can be locally approximated by a polynomial
        \begin{align}
            \hat{\memory}_\qbf(\kbf) = \Mbf^{[0]} + \Mbf^{[1]}(\kbf- \qbf) + \Mbf^{[2]}(\kbf - \qbf, \kbf - \qbf) + \ldots + \Mbf^{[p]}(\underbrace{\kbf - \qbf, \ldots, \kbf - \qbf}_{p\ \text{arguments}})\label{eqn:multivariate-polynomial}
        \end{align}
        where $\Mbf^{[j]}$ is an order $j$ multilinear map.
        Intuitively, $\Mcal^p$ is the class of functions that can be well approximated by an order $p$ Taylor expansion around each point.
        Consequently, the value retrieved from $\memory_{p,\, t}$ at a point $\qbf$ is given by the constant term of the polynomial: $\memory_{p}(\qbf) = \hat\memory_\qbf(\qbf) = \Mbf^{[0]}$.
    \item Choice of minimization algorithm: analytical solution.  Solving for the minimum is equivalent to solving for the set of tensors $(\Mbf^{[0]}, \ldots, \Mbf^{[p]})$ that minimize the least squares objective. Note that we need to jointly solve for all of $(\Mbf^{[0]}, \ldots, \Mbf^{[p]})$ to get the retrieved value $\Mbf^{[0]} = \memory_{p,\, t}(\qbf)$.
\end{enumerate}

Let us consider the simplest case of $p=0$, corresponding to the class of \emph{locally constant functions around $\qbf$}, also known as Nadaraya-Watson estimators \citep{Fan2018LocalPolynomialModelling,Zhang2024AnalysisAttentionLens,Goel2024CanTransformerRepresent}.
With this simplification, the output of our nonparametric associative memory is
\begin{align}
    m_{0,\, t}(\qbf) &= \argmin_{\Mbf^{[0]}\in \reals^\dkey} \frac{1}{2}\sum_{i=1}^t s(\kbf_i, \qbf) \norm{\vbf_i - \Mbf^{[0]}}_2^2 = \sum_{i=1}^t \frac{s(\kbf_i, \qbf)}{\sum_{j=1}^t s(\kbf_j, \qbf)} \vbf_i.\label{eqn:nadaraya-watson}
\end{align}
We may be disappointed at first that $s$ needs to be a monotonic function of the \emph{distance} $\norm{\kbf - \qbf}_2$ to enforce the local approximation property of the estimator in \autoref{eqn:local-polynomial-loss}; in contrast, softmax attention relies on an exponentiated \emph{inner product}.
However, if we normalize keys and queries such that $\norm{\kbf}_2=\norm{\qbf}_2=1$, the exponential smoothing kernel with bandwidth $B=2\sqrt{\dkey}$ is equivalent to the exponential scaled dot product:
\begin{align}
    s_{\text{exp}, 2\sqrt\dkey}(\kbf, \qbf) =\exp\left(\frac{-\norm{\kbf - \qbf}_2^2}{2\sqrt{\dkey}}\right) = \exp\left(\frac{-\norm{\kbf}_2^2 - \norm{\qbf}_2^2 + 2\kbf\transpose\qbf}{2\sqrt{\dkey}}\right) \propto \exp\left(\frac{\kbf\transpose\qbf}{\sqrt{\dkey}}\right),
\end{align}
allowing us to exactly derive softmax attention in \autoref{eqn:nadaraya-watson}.
Moreover, this preprocessing step lines up with an existing technique called query-key normalization (QKNorm), empirically known to stablize the training of large Transformers \citep{Dehghani2023ScalingVisionTransformers,Wortsman2023SmallscaleProxiesLargescale} but lacks theoretical justification.
Our derivation provides a simple explanation: \emph{softmax attention with QKNorm is a locally zeroth order non-parametric approximation of the true key-value map}.

By deriving self-attention from the same framework as the previous recurrent sequence layers, we can more clearly compare them through the perspective of how they do associative recall.
Previous sequence layers that we derived maintained an explicit matrix-valued recurrent state to compress the prefix key-value pairs, relying on this constant-sized matrix to perform retrieval.
In contrast, self-attention stores the entire set of prefix key-value pairs as its associative memory system, using all of them for retrieval.
As a result, its memory (and expressiveness) grows with the sequence length, a typical property of non-parametric methods \citep{Wang2019ExactGaussianProcesses,Wasserman2006AllNonparametricStatistics}.

\paragraph{Higher-order attention.}
\autoref{eqn:local-polynomial-loss} suggests a path to higher-order generalizations of self-attention by considering $p > 0$ at the cost of more computation.
By reducing \autoref{eqn:local-polynomial-loss} to a weighted least squares problem, we see the $p=1$ case of higher-order softmax self-attention is
\begin{align}
    \bmat{\Mbf^{[0]} & \Mbf^{[1]}} &= (\Vbf_t\transpose \Wbf_t \Xbf_t)(\Xbf_t\transpose \Wbf_t \Xbf_t)\inverse, &
    \Wbf_t &\triangleq \diag{s(\kbf_1, \qbf), \ldots, s(\kbf_t, \qbf)} \label{eqn:p1-softmax}\\
    \Xbf_t &= \bmat{1 & \ldots & 1\\ \kbf_1 - \qbf & \ldots & \kbf_t - \qbf}\transpose &
    \Vbf_t &= \bmat{\vbf_1 & \ldots & \vbf_t}\transpose,
\end{align}
where $\Wbf_t\in \reals^{t\times t},\ \Xbf_t \in \reals^{t\times (1 + \dkey)}$,\ $\Mbf^{[0]} \in \reals^{\dval}$, and $\Mbf^{[1]} \in \reals^{\dval \times \dkey}$.
These equations correspond to the standard prediction equations for local linear regression \citep{Fan2018LocalPolynomialModelling}.

An alternate form for the prediction $\Mbf^{[0]}$ that is more intuitive is
\begin{align}
    \Mbf^{[0]} &= \bar{\vbf} - \Mbf^{[1]}\bar{\dbf},&
    \bar{\vbf} &\triangleq \sum_{i=1}^t \frac{s(\kbf_i, \qbf)}{\sum_{j=1}^t s(\kbf_j, \qbf)} \vbf_i, &
    \bar{\dbf} &\triangleq \sum_{i=1}^t \frac{s(\kbf_i, \qbf)}{\sum_{j=1}^t s(\kbf_j, \qbf)} \dbf_i, &
    \dbf_i &\triangleq \kbf_i - \qbf,
\end{align}
resulting in an offset softmax attention as the output: $m_{1,\, t}(\qbf) = \Mbf^{[0]} = \bar{\vbf} - \Mbf^{[1]}\bar{\dbf}$.
This alternate form comes from the fact that if we already know $\Mbf^{[1]}$, then we can solve the $p=0$ case with an alternate regression problem with preprocessed data $\{(\kbf_i, \vbf_i - \Mbf^{[1]}\dbf_i)\}_{i=1}^t$, resulting in $\Mbf^{[0]}$ being a normalized weighted sum of $\vbf_i - \Mbf^{[1]}\dbf_i$.
Unlike standard softmax attention (the $p=0$ case), which only considers the interaction \emph{between keys and queries}, the $p=1$ case in \autoref{eqn:p1-softmax} also accounts for the covariance \emph{between keys} through the inverse term, similar to how linear regression improves upon linear attention from before.

Unfortunately, directly computing \autoref{eqn:p1-softmax} for $\qbf\in \{\qbf_1, \ldots, \qbf_T\}$ in parallel to fully leverage hardware-acceleration is impractical for all but the smallest sequence problems.
Materializing all $T$ prefix set of input tokens $\{\Xbf_t\}_{t=1}^T$ requires $\bigo{T^2\dkey}$ space, and materializing all $T$ prefix weighted sums of outer products $\{\Xbf_t\transpose\Wbf_t\Xbf_t\}_{t=1}^T$ requires $\bigo{T\dkey^2}$ space.
Computing $\{\Xbf\transpose_t\Wbf_t \allowbreak \Xbf_t\}_{t=1}^T$ requires $\bigo{T^2\dkey}$ time, and solving a batch of $T$ linear systems requires $\bigo{T\dkey^3}$ time.
These challenges makes the current form computationally impractical for practical sequence tasks.
Developing hardware-aware optimizations to compute \autoref{eqn:p1-softmax} similar to FlashAttention \citep{Dao2022FlashAttentionFastMemoryEfficient} is a promising future direction.

\section{Online regression with sequence layers}\label{sec:experiments}
We now empirically illustrate that the sequence layers discussed so far do in fact perform regression implicitly via a single forward pass.
We generate a sequence of keys according to a non-stationary, switching, autoregressive process:
\begin{align}
\kbf_{t+1} = 
\begin{cases}
0.9\kbf_{t} + 0.02 \bm{\varepsilon}_t,\, t < T/4 \\
0.999\kbf_{t} + 0.02 \bm{\varepsilon}_t,\, t \geq T/4
\end{cases}
\end{align}
with $\dkey=64$ and $T=256$.
Each $\bm{\varepsilon}_t \sim \Normal{0, \Sigmabf}$ where $\Sigmabf$ has unit diagonals and non-zero off-diagonals.
The choice of a non-identity $\Sigma$ creates instantaneous correlation between the dimensions of the keys so that $\Kbf_t\transpose\Kbf_t \neq \Ibf$.
We set the value vectors to be $\vbf_t = \kbf_{t+1}/\norm{\kbf_{t+1}}_2$, such that the dataset $\{(\kbf_i, \vbf_i)\}_{i=1}^T$ imitates a next-token prediction task.
We make one-step-ahead predictions by setting $\qbf_t = \kbf_{t+1}$ so that $\ybf_t = \memory_t(\qbf_t) = \memory_t(\kbf_{t+1})$ is the test-time regressor's out-of-sample prediction of the next timestep, conditioned on the prefix $\{(\kbf_i, \vbf_i)\}_{i=1}^t$.
To solve this online regression problem, we simply compute a single forward pass through each sequence layer.
\begin{figure}
  \begin{center}
    \includegraphics[width=0.5\linewidth]{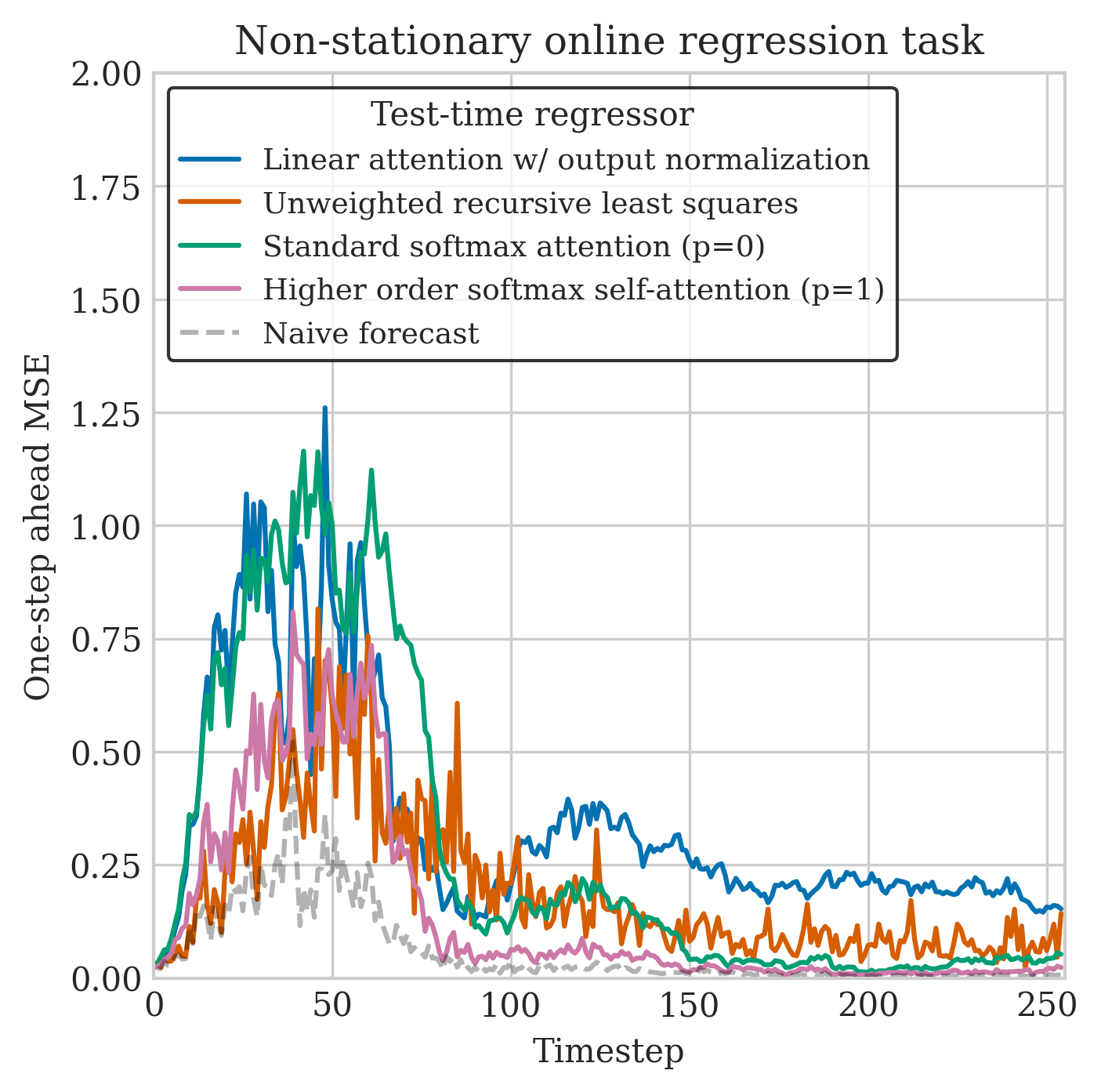}
  \end{center}
  \caption{Solving a non-stationary online regression task \emph{without any learnable parameters}. The regression-memory correspondence enables us to perform regression via a single forward pass of a test-time regression layer. The inputs change more rapidly in initial $T/4$ timesteps, making it harder for regressors to learn. The latter $3T/4$ timesteps are more stable; however, without decaying weights, linear attention and unweighted RLS are unable to adjust to the resulting non-stationary data.}
  \label{fig:nonstationary}
\end{figure}

We evaluate layers derived both from parametric regression and non-parametric regression.
For our parametric regression layers we choose linear attention (\autoref{eqn:linear-attention-recurrence}) and recursive least squares (\autoref{eqn:ols}) to show the effects of accounting for correlations between keys.
For our non-parametric regression layers, we chose softmax attention (\autoref{eqn:softmax-attention}) and our $p=1$ generalization of softmax attention based on local linear regression (\autoref{eqn:p1-softmax}).
We also show the naive one-step forecaster baseline where we simply use $\vbf_{t-1}$ as the prediction given $\kbf_t$.

We plot the one-step out-of-sample test loss $\norm{\vbf_{t+1} - \memory_t(\kbf_{t+1})}_2^2$ in \autoref{fig:nonstationary}.
During the initial $T/4$ timesteps, the dynamical system has a faster decay time with more rapid changes, making it less predictable.
In the latter $3T/4$ timesteps, the system slows down and is more predictable.
However, unweighted recursive least squares is unable to adapt to this change since it cannot discount datapoints from before the transition; as a result, it maintains a high error throughout the latter timesteps.
In contrast, the nonparametric regressors are able to adapt quickly to the transition since earlier keys are dissimilar to the later keys.
Interestingly, increasing our nonparametric softmax regressor from $p=0$ to $p=1$ allows the model to better adapt to both the initial fast-changing period and the sudden regime change.
As expected, linear attention performs the worst since $\Kbf_t\transpose\Kbf_t \neq \Ibf$ in our dynamical system\footnote{We used a standard output normalization for linear attention \citep{Qin2022DevilLinearTransformer} to prevent the output from scaling linearly with the sequence length.}.

\section{Constructing effective key-value pairs for next-token recall}\label{sec:key-construction}
Having shown that models derived via regression are able to perform associative recall, we now discuss the importance of constructing test-time key-value pairs that are pertinent to the task at hand.
This echoes an unchanging principle in machine learning: a well-designed model is only as effective as the data it processes.

Historically, query-key-value sequences were constructed via a linear projection on the corresponding input $\xbf_t$ at that timestep, e.g.  $\kbf_t = \Wbf_K\xbf_t$ \citep{Vaswani2017AttentionAllYou}.
Recent work on recurrent models, starting with \citet{Fu2022HungryHungryHippos}, identified the importance of performing a ``short convolution'' in addition to the per-timestep projection:
\begin{align}
    \kbf_t = \sum_{i=0}^{K-1} w_i \Wbf_K \xbf_{t-i} \label{eqn:short-conv}
\end{align}
where $K$ is the length of the short convolution filter.
The short convolution imitates the induction heads behavior learned by self-attention layers near the start of the neural network,  which allow Transformers to ``look back'' at previous tokens.
Setting $w_0=i$ and $w_j=0$ for $j\neq 0$ in \autoref{eqn:short-conv} enables other sequence layers to also ``look back'' at the token from timestep $i$.

The short convolution is a crucial component of purely recurrent language models; removing the short convolution results in severe performance drops \citep{Yang2024GatedLinearAttention,Yang2024ParallelizingLinearTransformers,Yang2024GatedDeltaNetworks,Sun2024LearningLearnTest}.
Including a short convolution before each self-attention layer can even improve the performance of Transformers, by freeing the earlier self-attention layers from having to learn the induction head behavior \citep{Xu2024KVShiftingAttention}.
In this section, we provide a partial explanation for why this short convolution is so crucial, through our perspective of associative memory.
Specifically, we show that this short convolution allows the sequence layer to memorize and retrieve bigram-like key-value pairs, which has been shown to be important for language modeling \citep{Arora2023ZoologyMeasuringImproving}.

\paragraph{One short conv is all you need for next-token recall.}
Consider a standard associative recall task introduced by \citet{Arora2023ZoologyMeasuringImproving}, in which a model must be able to recall multiple key-value pairings, known as multi-query associative recall (MQAR).
In this task, there are $P$ pairs of unique and consistent cue-response\footnote{We use the ``cue'' and ``response'' terminology to avoid confusion with ``keys'' and ``values'' in the context of test-time regression.} tokens $\{(\ubf_j, \vbf_j)\}_{j\in [P]}$ with each cue $\ubf_j$ mapping one-to-one to its response $\vbf_j$.
The model is given a contextual sequence of randomly drawn, possibly repeated, cue-response pairs followed by a previously seen cue: $(\xbf_1, \xbf_2, \ldots, \xbf_{T-1}, \xbf_T, \xbf_{T+1}) = (\ubf_{i_1}, \vbf_{i_1}, \ldots, \ubf_{i_{T/2}}, \vbf_{i_{T/2}}, \ubf_{i_j})$. $\ubf_{i_j}$ is a cue that appeared earlier (the $j$th pair) and the model is expected to output the paired response $\vbf_{i_j}$ at timestep $T+1$.
This benchmark abstracts the ``Hakuna Matata`` example ``Hakuna Matata, it means no worries for the rest of your days. Hakuna \underline{\hspace{1cm}} '' from \citet{Arora2023ZoologyMeasuringImproving}, which we also discussed in \autoref{sec:introduction}.

We now show that a single test-time regression layer (e.g. as simple as linear attention) with one short convolution is sufficient to solve MQAR, without any parameters other than the embeddings, \emph{as long as it is provided with the appropriate key-value pairs to regress}.

To solve this task, it suffices to test-time memorize all the bigram pairing $\{(\ubf_j, \vbf_j)\}$.
Given $\xbf_1, \ldots, \xbf_T$, we causally construct our test-time regression dataset as follows.
Let the keys be constructed via a short convolution $\kbf_t = w_0 \xbf_t + w_1 \xbf_{t-1} = \xbf_{t-1}$ where $w_0 = 0, w_1 = 1$.
Let the queries and values be the same as the inputs: $\qbf_t = \vbf_t = \xbf_t$.
In other words, our AR sequence layer will first memorize $\{(\kbf_t, \vbf_t)\}_{t=1}^{T+1} = \{(\xbf_{t-1}, \xbf_t)\}_{t=1}^{T+1}$ into its associative memory map $\memory_{T+1}$.
Then we retrieve from $\memory_{T+1}$ using $\qbf_{T+1} = \xbf_{T+1} = \ubf_{i_j}$, which has previously appeared in the sequence of cues as the $i_j$th bigram.
The output is then
\begin{align}
\ybf_{T+1} = \memory_{T+1}(\qbf_{T+1}) = \memory_{T+1}(\ubf_{i_j}).    
\end{align}
For simplicity, suppose $\memory_{T+1}$ is implemented via linear attention, following \autoref{eqn:linear-attention-recurrence}.
Then
\begin{align}
    \ybf_{T+1} = \memory_{T+1}(\ubf_{i_j}) = \left(\sum_{t=1}^{T+1}\vbf_t\kbf_t\transpose\right) \ubf_{i_j} = \sum_{t=1}^{T+1} \xbf_t \xbf_{t-1}\transpose \ubf_{i_j}. \label{eqn:solve-mqar}
\end{align}
When the embedding space is large enough, we can set the embeddings such that all tokens are orthonormal.
Then $\xbf_{t-1}\transpose \ubf_{i_j}$ is nonzero only when $\xbf_{t-1} = \ubf_{i_j}$, simplifying the output to a sum of only tokens that are followed by $\ubf_{i_j}$.
Since the cue-response map is one-to-one, the remaining tokens are all $\vbf_{i_j}$, producing the output $\ybf_{T+1} \propto \vbf_{i_j}$, solving the MQAR task.

\paragraph{Memory capacity, not sequence length, limits model performance.}
\begin{figure}
  \begin{center}
    \includegraphics[width=0.5\linewidth]{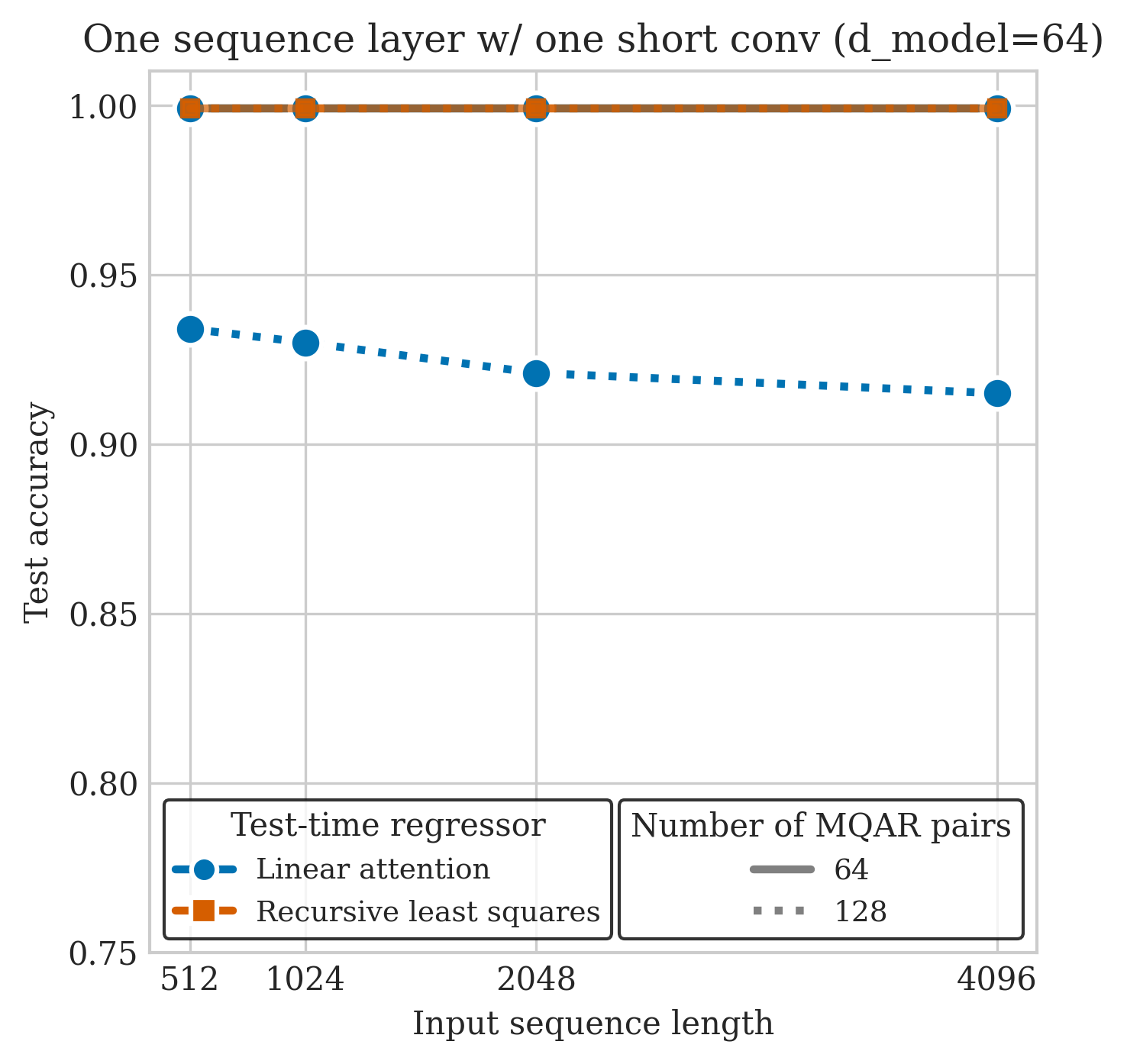}
  \end{center}
  \caption{A single test-time regression layer with one short convolution suffices to solve MQAR with $P=64$ cue-response pairs. When a model has sufficient capacity to memorize all $P$ pairs, it can solve MQAR \emph{regardless} of the sequence length. The regression problem is stationary (the cue-response map doesn't change over time), hence no forgetting mechanism is needed to solve the task.}
  \label{fig:mqar}
\end{figure}

Typical evaluations of models on the MQAR task look at model performance with respect to the length of the sequence, possibly varying the model capacity. A few examples include Figure 2 of \citet{Arora2023ZoologyMeasuringImproving}, Figure 8 of \citet{Dao2024TransformersAreSSMs}, Figure 2 of \citet{Yang2024ParallelizingLinearTransformers}, and likely others.
However, our above construction shows that the sequence length doesn't matter for memorizers with a large enough memory capacity; once the memory is large enough, the difficulty of MQAR is \emph{independent} of the sequence length $T$, which may be counter to the intuitive notion of longer sequences being more difficult.
Instead, models are \emph{limited by their ability to fully memorize all $P$ key-value pairs}.

Although our construction in \autoref{eqn:solve-mqar} relied on the embedding space being large enough to allow orthonormality, trained models can still succeed without perfectly orthonormal embeddings, which we empirically show in \autoref{fig:mqar}.
We train on the MQAR task following the procedure of \citep{Arora2023ZoologyMeasuringImproving}.
However, \emph{we use only one test-time regression layer and one length-2 convolution} (for constructing the keys), following our construction.
We also remove the typical MLP block following the sequence layer, since our construction shows that additional nonlinearities are unnecessary for this synthetic associative recall task.
We fix our key dimensions $\dkey$ to be the same as the embedding size, $d_{\text{model}} = 64$.
We compare two test-time regression architectures, linear attention and unweighted recursive least squares (RLS) of \autoref{eqn:ols}.
We do not use gated variants of linear attention since this task does not require any forgetting; the cue-response mapping is unchanged throughout the sequence.
As shown by our previous construction, even linear attention is able to solve the task perfectly when the appropriate keys are constructed.
In the case where $d_{\text{model}} = P=64$, linear attention solves MQAR perfectly, independent of sequence length as predicted.
Once the number of keys increases to $P=128$, linear attention can no longer perfectly solve MQAR, since it is impossible for 128 keys to be orthogonal in a $d_{\text{model}}=64$ dimensional space.
In contrast, an unweighted RLS layer (\autoref{eqn:ols}) improves upon linear attention by accounting for the key covariance matrices $\{\Kbf_t\transpose\Kbf_t\}$, allowing it to better test-time memorize the cue-response map, even when $P > d_{\text{model}}$.
As a result, RLS is able to solve MQAR regardless of the sequence length.

\section{Related works}\label{sec:related-works}
Our test-time regression framework was derived by formalizing the notion of associative recall, resulting in sequence layers that first memorize then retrieve.
Designing neural networks motivated with associative memory has a long history, dating back as early as the works of \citet{Hopfield1982NeuralNetworksPhysical}, \citet{Willshaw1969NonHolographicAssociativeMemory},\citet{Willshaw1989HolographyAssociativeMemory},\citet{Kohonen1972CorrelationMatrixMemories}, and \citet{Hinton1989ParallelModelsAssociative}.
Many more contemporary works have also sought to understand and derive neural network layers from the perspective of memory, drawing inspiration from both neuroscience and dynamical systems \citep{Bietti2023BirthTransformerMemory,Krotov2016DenseAssociativeMemory,Krotov2021LargeAssociativeMemory, Ramsauer2020HopfieldNetworksAll,Millidge2022UniversalHopfieldNetworks,Behrouz2024TitansLearningMemorize}.
Indeed, even the feedforward layers of an MLP can be seen as a persistent associative memory \citep{Sukhbaatar2019AugmentingSelfattentionPersistent,Zhang2025MemoryMosaics,Bietti2023BirthTransformerMemory,Nichani2024UnderstandingFactualRecall,Cabannes2023ScalingLawsAssociative}.

The ability to reference and retrieve past tokens for further computation is believed to be key to in-context learning and zero-shot/few-shot learning in large language models \citep{Olsson2022ContextLearningInduction,Dong2024SurveyIncontextLearning}.
There have been many attempts at understanding how sequence models are able to learn in-context \citep{Garg2022WhatCanTransformers,Zhang2023WhatHowDoes}.
Some works have shown that Transformers, due to their self-attention layers, learn linear functions in-context by learning to perform gradient descent, where the iterates are updated \emph{per layer} \citep{Oswald2023TransformersLearnInContext,Ahn2023TransformersLearnImplement}.
This phenomenon is sometimes known as mesa-optimization \citep{vonOswald2023UncoveringMesaoptimizationAlgorithms} where the model learns to solve an auxiliary optimization problem in order to minimize its training objective.
However, sometimes these intermediate optimization processes are actually implicitly in the architecture itself, as we showed in \autoref{eqn:linear-attention-gd} where each linear attention layer (and its gated variant) explicitly performs one step of gradient descent. 
Note that this behavior is different from the class of recurrent models that perform stochastic gradient descent \emph{within} the layer, as defined by \autoref{eqn:sgd}.

\section{Conclusion and discussion}\label{sec:discussion}

We have presented a unifying framework that derives sequence models that can explicitly perform associative recall via test-time regression. Each architecture emerges naturally by specifying three aspects: the relative importance of each association, the function class, and the optimization algorithm. Through this lens, we derived linear attention and its variants as approximations to the linear least squares solution, online learning layers and fast-weight layers as first-order least-squares solvers, and softmax attention as a local constant regressor.
This perspective provided principled explanations for architectural choices like QKNorm and derived higher-order generalizations of attention that go beyond pairwise interactions.
We also demonstrated that for multi-query associative recall (MQAR), a single short convolution and a sequence layer suffices, in contrast to typical approaches that use multiple layers and unnecessary feedforward blocks \citep{Arora2023ZoologyMeasuringImproving}.

Our paper focuses specifically on sequence architectures that use query-key-value patterns for associative recall.
Several important architectural families lie outside this scope, including structured state-space models with more general structured masks \citep{Dao2024TransformersAreSSMs}, convolutional architectures \citep{Shi2023SequenceModelingMultiresolutiona,Poli2023HyenaHierarchyLarger,Gehring2017ConvolutionalSequenceSequence,Gu2021EfficientlyModelingLong,Bradbury2017QuasirecurrentNeuralNetworks,Smith2023SimplifiedStateSpace,Hochreiter1997LongShortTermMemory}, and more novel computational patterns \citep{Qin2023ToeplitzNeuralNetwork,Ren2023SparseModularActivation}. Moreover, complementary research on model backbones and initialization continues to yield important advances and performance improvements \citep{Gu2024MambaLinearTimeSequence,Dao2024TransformersAreSSMs,Orvieto2023ResurrectingRecurrentNeural,Mehta2023LongRangeLanguage,Hua2022TransformerQualityLinear}.

Our framework opens rich directions for future research by connecting to the extensive literature on regression and optimization. While initial successes like TTT-MLP \citep{Sun2024LearningLearnTest} hint at the potential of neural test-time regressors, the space of nonlinear neural regression models remains largely unexplored.
Recent work integrating weight decay and momentum \citep{Behrouz2024TitansLearningMemorize,Yang2024GatedDeltaNetworks} suggests opportunities for developing more effective, stable, and parallelizable test-time optimizers.
Concurrent work by \citet{Behrouz2024TitansLearningMemorize} also suggests opportunities to more creatively utilize associative memory.
The practical success of these approaches will depend heavily on efficient hardware implementations \citep{Hua2022TransformerQualityLinear,Yang2024ParallelizingLinearTransformers,Sun2024LearningLearnTest,Dao2024TransformersAreSSMs,Dao2022FlashAttentionFastMemoryEfficient},
aligning with established compute-scaling principles \citep{Sutton2019BitterLesson,Hoffmann2022TrainingComputeOptimalLarge,Kaplan2020ScalingLawsNeural}.

In the words of \citet{Kohonen1989SelfOrganizationAssociativeMemory}, ``associative memory is a very delicate and complex concept which often
has been attributed to the higher cognitive processes, especially those taking
place in the human brain''.
We speculate that test-time associative memory, along with test-time compute more broadly \citep{Sun2024LearningLearnTest, Sun2020TestTimeTrainingSelfSupervision,Akyurek2024SurprisingEffectivenessTestTime}, will be fundamental to developing truly adaptive models that can update and learn in changing environments \citep{OpenAI2024OpenAIO1System}.

\paragraph{Acknowledgements.} This work was supported in part by AFOSR Grant FA9550-21-1-0397, ONR Grant N00014-22-1-2110, NSF Grant 2205084, the Stanford Institute for Human-Centered Artificial Intelligence (HAI). EBF is a Chan Zuckerberg Biohub – San Francisco Investigator.
{\small
    \bibliography{references}

\begin{thebibliography}{120}
\providecommand{\natexlab}[1]{#1}
\providecommand{\url}[1]{\texttt{#1}}
\expandafter\ifx\csname urlstyle\endcsname\relax
  \providecommand{\doi}[1]{doi: #1}\else
  \providecommand{\doi}{doi: \begingroup \urlstyle{rm}\Url}\fi

\bibitem[Ahn et~al.(2023)Ahn, Cheng, Daneshmand, and Sra]{Ahn2023TransformersLearnImplement}
K.~Ahn, X.~Cheng, H.~Daneshmand, and S.~Sra.
\newblock Transformers learn to implement preconditioned gradient descent for in-context learning.
\newblock \emph{Advances in Neural Information Processing Systems}, 36:\penalty0 45614--45650, Dec. 2023.
\newblock URL \url{https://proceedings.neurips.cc/paper_files/paper/2023/hash/8ed3d610ea4b68e7afb30ea7d01422c6-Abstract-Conference.html}.

\bibitem[Ainslie et~al.(2023)Ainslie, Lee-Thorp, Jong, Zemlyanskiy, Lebron, and Sanghai]{Ainslie2023GQATrainingGeneralized}
J.~Ainslie, J.~Lee-Thorp, M.~d. Jong, Y.~Zemlyanskiy, F.~Lebron, and S.~Sanghai.
\newblock {GQA}: {Training} {Generalized} {Multi}-{Query} {Transformer} {Models} from {Multi}-{Head} {Checkpoints}.
\newblock In \emph{The 2023 {Conference} on {Empirical} {Methods} in {Natural} {Language} {Processing}}, Dec. 2023.
\newblock URL \url{https://openreview.net/forum?id=hmOwOZWzYE}.

\bibitem[Aksenov et~al.(2024)Aksenov, Balagansky, Vaina, Shaposhnikov, Gorbatovski, and Gavrilov]{Aksenov2024LinearTransformersLearnable}
Y.~Aksenov, N.~Balagansky, S.~M. L.~C. Vaina, B.~Shaposhnikov, A.~Gorbatovski, and D.~Gavrilov.
\newblock Linear {Transformers} with {Learnable} {Kernel} {Functions} are {Better} {In}-{Context} {Models}, Feb. 2024.
\newblock URL \url{http://arxiv.org/abs/2402.10644}.
\newblock arXiv:2402.10644 [cs] version: 1.

\bibitem[Akyürek et~al.(2024)Akyürek, Damani, Qiu, Guo, Kim, and Andreas]{Akyurek2024SurprisingEffectivenessTestTime}
E.~Akyürek, M.~Damani, L.~Qiu, H.~Guo, Y.~Kim, and J.~Andreas.
\newblock The {Surprising} {Effectiveness} of {Test}-{Time} {Training} for {Abstract} {Reasoning}, Nov. 2024.
\newblock URL \url{http://arxiv.org/abs/2411.07279}.
\newblock arXiv:2411.07279 [cs] version: 1.

\bibitem[Ansari et~al.(2024)Ansari, Stella, Turkmen, Zhang, Mercado, Shen, Shchur, Rangapuram, Arango, Kapoor, Zschiegner, Maddix, Wang, Mahoney, Torkkola, Wilson, Bohlke-Schneider, and Wang]{Ansari2024ChronosLearningLanguage}
A.~F. Ansari, L.~Stella, A.~C. Turkmen, X.~Zhang, P.~Mercado, H.~Shen, O.~Shchur, S.~S. Rangapuram, S.~P. Arango, S.~Kapoor, J.~Zschiegner, D.~C. Maddix, H.~Wang, M.~W. Mahoney, K.~Torkkola, A.~G. Wilson, M.~Bohlke-Schneider, and B.~Wang.
\newblock Chronos: {Learning} the language of time series.
\newblock \emph{Transactions on Machine Learning Research}, 2024.
\newblock ISSN 2835-8856.
\newblock URL \url{https://openreview.net/forum?id=gerNCVqqtR}.

\bibitem[Arora et~al.(2023{\natexlab{a}})Arora, Eyuboglu, Timalsina, Johnson, Poli, Zou, Rudra, and Re]{Arora2023ZoologyMeasuringImproving}
S.~Arora, S.~Eyuboglu, A.~Timalsina, I.~Johnson, M.~Poli, J.~Zou, A.~Rudra, and C.~Re.
\newblock Zoology: {Measuring} and {Improving} {Recall} in {Efficient} {Language} {Models}.
\newblock In \emph{The {Twelfth} {International} {Conference} on {Learning} {Representations}}, Oct. 2023{\natexlab{a}}.
\newblock URL \url{https://openreview.net/forum?id=LY3ukUANko}.

\bibitem[Arora et~al.(2023{\natexlab{b}})Arora, Yang, Eyuboglu, Narayan, Hojel, Trummer, and Ré]{Arora2023LanguageModelsEnable}
S.~Arora, B.~Yang, S.~Eyuboglu, A.~Narayan, A.~Hojel, I.~Trummer, and C.~Ré.
\newblock Language {Models} {Enable} {Simple} {Systems} for {Generating} {Structured} {Views} of {Heterogeneous} {Data} {Lakes}, Apr. 2023{\natexlab{b}}.
\newblock URL \url{http://arxiv.org/abs/2304.09433}.
\newblock arXiv:2304.09433 [cs].

\bibitem[Arora et~al.(2024)Arora, Eyuboglu, Zhang, Timalsina, Alberti, Zou, Rudra, and Re]{Arora2024SimpleLinearAttention}
S.~Arora, S.~Eyuboglu, M.~Zhang, A.~Timalsina, S.~Alberti, J.~Zou, A.~Rudra, and C.~Re.
\newblock Simple linear attention language models balance the recall-throughput tradeoff.
\newblock In \emph{Proceedings of the 41st {International} {Conference} on {Machine} {Learning}}, pages 1763--1840. PMLR, July 2024.
\newblock URL \url{https://proceedings.mlr.press/v235/arora24a.html}.
\newblock ISSN: 2640-3498.

\bibitem[Beck et~al.(2024)Beck, Pöppel, Spanring, Auer, Prudnikova, Kopp, Klambauer, Brandstetter, and Hochreiter]{Beck2024XLSTMExtendedLong}
M.~Beck, K.~Pöppel, M.~Spanring, A.~Auer, O.~Prudnikova, M.~Kopp, G.~Klambauer, J.~Brandstetter, and S.~Hochreiter.
\newblock {xLSTM}: {Extended} {Long} {Short}-{Term} {Memory}, May 2024.
\newblock URL \url{http://arxiv.org/abs/2405.04517}.
\newblock arXiv:2405.04517 [cs, stat].

\bibitem[Behrouz et~al.(2024)Behrouz, Zhong, and Mirrokni]{Behrouz2024TitansLearningMemorize}
A.~Behrouz, P.~Zhong, and V.~Mirrokni.
\newblock Titans: {Learning} to {Memorize} at {Test} {Time}, Dec. 2024.
\newblock URL \url{http://arxiv.org/abs/2501.00663}.
\newblock arXiv:2501.00663 [cs].

\bibitem[Bertasius et~al.(2021)Bertasius, Wang, and Torresani]{Bertasius2021SpaceTimeAttentionAll}
G.~Bertasius, H.~Wang, and L.~Torresani.
\newblock Is {Space}-{Time} {Attention} {All} {You} {Need} for {Video} {Understanding}?
\newblock In \emph{Proceedings of the 38th {International} {Conference} on {Machine} {Learning}}, pages 813--824. PMLR, July 2021.
\newblock URL \url{https://proceedings.mlr.press/v139/bertasius21a.html}.
\newblock ISSN: 2640-3498.

\bibitem[Bietti et~al.(2023)Bietti, Cabannes, Bouchacourt, Jegou, and Bottou]{Bietti2023BirthTransformerMemory}
A.~Bietti, V.~Cabannes, D.~Bouchacourt, H.~Jegou, and L.~Bottou.
\newblock Birth of a {Transformer}: {A} {Memory} {Viewpoint}.
\newblock In \emph{Thirty-seventh {Conference} on {Neural} {Information} {Processing} {Systems}}, Nov. 2023.
\newblock URL \url{https://openreview.net/forum?id=3X2EbBLNsk}.

\bibitem[Boyd and Vandenberghe(2004)]{Boyd2004ConvexOptimization}
S.~Boyd and L.~Vandenberghe.
\newblock \emph{Convex optimization}.
\newblock Cambridge University Press, Cambridge, 2004.

\bibitem[Bradbury et~al.(2017)Bradbury, Merity, Xiong, and Socher]{Bradbury2017QuasirecurrentNeuralNetworks}
J.~Bradbury, S.~Merity, C.~Xiong, and R.~Socher.
\newblock Quasi-recurrent neural networks.
\newblock In \emph{International conference on learning representations}, 2017.
\newblock URL \url{https://openreview.net/forum?id=H1zJ-v5xl}.

\bibitem[Brown et~al.(2020)Brown, Mann, Ryder, Subbiah, Kaplan, Dhariwal, Neelakantan, Shyam, Sastry, Askell, Agarwal, Herbert-Voss, Krueger, Henighan, Child, Ramesh, Ziegler, Wu, Winter, Hesse, Chen, Sigler, Litwin, Gray, Chess, Clark, Berner, McCandlish, Radford, Sutskever, and Amodei]{Brown2020LanguageModelsAre}
T.~B. Brown, B.~Mann, N.~Ryder, M.~Subbiah, J.~Kaplan, P.~Dhariwal, A.~Neelakantan, P.~Shyam, G.~Sastry, A.~Askell, S.~Agarwal, A.~Herbert-Voss, G.~Krueger, T.~Henighan, R.~Child, A.~Ramesh, D.~M. Ziegler, J.~Wu, C.~Winter, C.~Hesse, M.~Chen, E.~Sigler, M.~Litwin, S.~Gray, B.~Chess, J.~Clark, C.~Berner, S.~McCandlish, A.~Radford, I.~Sutskever, and D.~Amodei.
\newblock Language {Models} are {Few}-{Shot} {Learners}, July 2020.
\newblock URL \url{http://arxiv.org/abs/2005.14165}.
\newblock arXiv:2005.14165 [cs].

\bibitem[Cabannes et~al.(2023)Cabannes, Dohmatob, and Bietti]{Cabannes2023ScalingLawsAssociative}
V.~Cabannes, E.~Dohmatob, and A.~Bietti.
\newblock Scaling {Laws} for {Associative} {Memories}.
\newblock In \emph{The {Twelfth} {International} {Conference} on {Learning} {Representations}}, Oct. 2023.
\newblock URL \url{https://openreview.net/forum?id=Tzh6xAJSll}.

\bibitem[Castoldi and de~Campos(2009)]{Castoldi2009MinimumdisturbanceDescriptionDevelopment}
F.~T. Castoldi and M.~L.~R. de~Campos.
\newblock Minimum-disturbance description for the development of adaptation algorithms and a new leakage least squares algorithm.
\newblock In \emph{2009 {IEEE} {International} {Conference} on {Acoustics}, {Speech} and {Signal} {Processing}}, pages 3129--3132, Apr. 2009.
\newblock \doi{10.1109/ICASSP.2009.4960287}.
\newblock URL \url{https://ieeexplore.ieee.org/document/4960287}.
\newblock ISSN: 2379-190X.

\bibitem[Cesa-Bianchi et~al.(2004)Cesa-Bianchi, Conconi, and Gentile]{Cesa-Bianchi2004GeneralizationAbilityOnline}
N.~Cesa-Bianchi, A.~Conconi, and C.~Gentile.
\newblock On the generalization ability of on-line learning algorithms.
\newblock \emph{IEEE Transactions on Information Theory}, 50\penalty0 (9):\penalty0 2050--2057, Sept. 2004.
\newblock ISSN 1557-9654.
\newblock \doi{10.1109/TIT.2004.833339}.
\newblock URL \url{https://ieeexplore.ieee.org/document/1327806/?arnumber=1327806}.
\newblock Conference Name: IEEE Transactions on Information Theory.

\bibitem[Chen et~al.(2024)Chen, Liuzhicheng, Wang, Tian, and Wang]{Chen2024DiJiangEfficientLarge}
H.~Chen, Liuzhicheng, X.~Wang, Y.~Tian, and Y.~Wang.
\newblock {DiJiang}: {Efficient} {Large} {Language} {Models} through {Compact} {Kernelization}.
\newblock In \emph{Forty-first {International} {Conference} on {Machine} {Learning}}, June 2024.
\newblock URL \url{https://openreview.net/forum?id=0uUHfhXdnH&referrer=%5Bthe%20profile%20of%20Yuchuan%20Tian%5D(%2Fprofile%3Fid%3D~Yuchuan_Tian1)}.

\bibitem[Chen et~al.(2021)Chen, Zeng, Ji, and Yang]{Chen2021SkyformerRemodelSelfattention}
Y.~Chen, Q.~Zeng, H.~Ji, and Y.~Yang.
\newblock Skyformer: remodel self-attention with {Gaussian} kernel and nyström method.
\newblock In \emph{Proceedings of the 35th {International} {Conference} on {Neural} {Information} {Processing} {Systems}}, {NIPS} '21, pages 2122--2135, Red Hook, NY, USA, 2021. Curran Associates Inc.
\newblock ISBN 978-1-7138-4539-3.

\bibitem[Choromanski et~al.(2020)Choromanski, Likhosherstov, Dohan, Song, Gane, Sarlos, Hawkins, Davis, Mohiuddin, Kaiser, Belanger, Colwell, and Weller]{Choromanski2020RethinkingAttentionPerformers}
K.~M. Choromanski, V.~Likhosherstov, D.~Dohan, X.~Song, A.~Gane, T.~Sarlos, P.~Hawkins, J.~Q. Davis, A.~Mohiuddin, L.~Kaiser, D.~B. Belanger, L.~J. Colwell, and A.~Weller.
\newblock Rethinking {Attention} with {Performers}.
\newblock In \emph{International {Conference} on {Learning} {Representations}}, Oct. 2020.
\newblock URL \url{https://openreview.net/forum?id=Ua6zuk0WRH}.

\bibitem[Clark et~al.(2022)Clark, Guu, Chang, Pasupat, Hinton, and Norouzi]{Clark2022MetaLearningFastWeight}
K.~Clark, K.~Guu, M.-W. Chang, P.~Pasupat, G.~Hinton, and M.~Norouzi.
\newblock Meta-{Learning} {Fast} {Weight} {Language} {Models}.
\newblock In Y.~Goldberg, Z.~Kozareva, and Y.~Zhang, editors, \emph{Proceedings of the 2022 {Conference} on {Empirical} {Methods} in {Natural} {Language} {Processing}}, pages 9751--9757, Abu Dhabi, United Arab Emirates, Dec. 2022. Association for Computational Linguistics.
\newblock \doi{10.18653/v1/2022.emnlp-main.661}.
\newblock URL \url{https://aclanthology.org/2022.emnlp-main.661}.

\bibitem[Dao and Gu(2024)]{Dao2024TransformersAreSSMs}
T.~Dao and A.~Gu.
\newblock Transformers are {SSMs}: {Generalized} {Models} and {Efficient} {Algorithms} {Through} {Structured} {State} {Space} {Duality}.
\newblock In \emph{Forty-first {International} {Conference} on {Machine} {Learning}}, June 2024.

\bibitem[Dao et~al.(2022)Dao, Fu, Ermon, Rudra, and Re]{Dao2022FlashAttentionFastMemoryEfficient}
T.~Dao, D.~Y. Fu, S.~Ermon, A.~Rudra, and C.~Re.
\newblock {FlashAttention}: {Fast} and {Memory}-{Efficient} {Exact} {Attention} with {IO}-{Awareness}.
\newblock In \emph{Advances in {Neural} {Information} {Processing} {Systems}}, Oct. 2022.
\newblock URL \url{https://openreview.net/forum?id=H4DqfPSibmx}.

\bibitem[De et~al.(2024)De, Smith, Fernando, Botev, Cristian-Muraru, Gu, Haroun, Berrada, Chen, Srinivasan, Desjardins, Doucet, Budden, Teh, Pascanu, Freitas, and Gulcehre]{De2024GriffinMixingGated}
S.~De, S.~L. Smith, A.~Fernando, A.~Botev, G.~Cristian-Muraru, A.~Gu, R.~Haroun, L.~Berrada, Y.~Chen, S.~Srinivasan, G.~Desjardins, A.~Doucet, D.~Budden, Y.~W. Teh, R.~Pascanu, N.~D. Freitas, and C.~Gulcehre.
\newblock Griffin: {Mixing} {Gated} {Linear} {Recurrences} with {Local} {Attention} for {Efficient} {Language} {Models}, Feb. 2024.
\newblock URL \url{http://arxiv.org/abs/2402.19427}.
\newblock arXiv:2402.19427 [cs].

\bibitem[Dehghani et~al.(2023)Dehghani, Djolonga, Mustafa, Padlewski, Heek, Gilmer, Steiner, Caron, Geirhos, Alabdulmohsin, Jenatton, Beyer, Tschannen, Arnab, Wang, Ruiz, Minderer, Puigcerver, Evci, Kumar, Steenkiste, Elsayed, Mahendran, Yu, Oliver, Huot, Bastings, Collier, Gritsenko, Birodkar, Vasconcelos, Tay, Mensink, Kolesnikov, Pavetic, Tran, Kipf, Lucic, Zhai, Keysers, Harmsen, and Houlsby]{Dehghani2023ScalingVisionTransformers}
M.~Dehghani, J.~Djolonga, B.~Mustafa, P.~Padlewski, J.~Heek, J.~Gilmer, A.~P. Steiner, M.~Caron, R.~Geirhos, I.~Alabdulmohsin, R.~Jenatton, L.~Beyer, M.~Tschannen, A.~Arnab, X.~Wang, C.~R. Ruiz, M.~Minderer, J.~Puigcerver, U.~Evci, M.~Kumar, S.~V. Steenkiste, G.~F. Elsayed, A.~Mahendran, F.~Yu, A.~Oliver, F.~Huot, J.~Bastings, M.~Collier, A.~A. Gritsenko, V.~Birodkar, C.~N. Vasconcelos, Y.~Tay, T.~Mensink, A.~Kolesnikov, F.~Pavetic, D.~Tran, T.~Kipf, M.~Lucic, X.~Zhai, D.~Keysers, J.~J. Harmsen, and N.~Houlsby.
\newblock Scaling {Vision} {Transformers} to 22 {Billion} {Parameters}.
\newblock In \emph{Proceedings of the 40th {International} {Conference} on {Machine} {Learning}}, pages 7480--7512. PMLR, July 2023.
\newblock URL \url{https://proceedings.mlr.press/v202/dehghani23a.html}.
\newblock ISSN: 2640-3498.

\bibitem[Devlin et~al.(2019)Devlin, Chang, Lee, and Toutanova]{Devlin2019BERTPretrainingDeep}
J.~Devlin, M.-W. Chang, K.~Lee, and K.~Toutanova.
\newblock {BERT}: {Pre}-training of {Deep} {Bidirectional} {Transformers} for {Language} {Understanding}.
\newblock In J.~Burstein, C.~Doran, and T.~Solorio, editors, \emph{Proceedings of the 2019 {Conference} of the {North} {American} {Chapter} of the {Association} for {Computational} {Linguistics}: {Human} {Language} {Technologies}, {Volume} 1 ({Long} and {Short} {Papers})}, pages 4171--4186, Minneapolis, Minnesota, June 2019. Association for Computational Linguistics.
\newblock \doi{10.18653/v1/N19-1423}.
\newblock URL \url{https://aclanthology.org/N19-1423/}.

\bibitem[Dong et~al.(2024)Dong, Li, Dai, Zheng, Ma, Li, Xia, Xu, Wu, Liu, Chang, Sun, Li, and Sui]{Dong2024SurveyIncontextLearning}
Q.~Dong, L.~Li, D.~Dai, C.~Zheng, J.~Ma, R.~Li, H.~Xia, J.~Xu, Z.~Wu, T.~Liu, B.~Chang, X.~Sun, L.~Li, and Z.~Sui.
\newblock A {Survey} on {In}-context {Learning}, Oct. 2024.
\newblock URL \url{http://arxiv.org/abs/2301.00234}.
\newblock arXiv:2301.00234 [cs].

\bibitem[Dosovitskiy et~al.(2021)Dosovitskiy, Beyer, Kolesnikov, Weissenborn, Zhai, Unterthiner, Dehghani, Minderer, Heigold, Gelly, Uszkoreit, and Houlsby]{Dosovitskiy2021ImageWorth16x16}
A.~Dosovitskiy, L.~Beyer, A.~Kolesnikov, D.~Weissenborn, X.~Zhai, T.~Unterthiner, M.~Dehghani, M.~Minderer, G.~Heigold, S.~Gelly, J.~Uszkoreit, and N.~Houlsby.
\newblock An image is worth 16x16 words: {Transformers} for image recognition at scale.
\newblock In \emph{International conference on learning representations}, 2021.
\newblock URL \url{https://openreview.net/forum?id=YicbFdNTTy}.

\bibitem[Duchi et~al.(2011)Duchi, Hazan, and Singer]{Duchi2011AdaptiveSubgradientMethods}
J.~Duchi, E.~Hazan, and Y.~Singer.
\newblock Adaptive subgradient methods for online learning and stochastic optimization.
\newblock \emph{Journal of Machine Learning Research}, 12\penalty0 (61):\penalty0 2121--2159, 2011.
\newblock URL \url{http://jmlr.org/papers/v12/duchi11a.html}.

\bibitem[Fan(2018)]{Fan2018LocalPolynomialModelling}
J.~Fan.
\newblock \emph{Local {Polynomial} {Modelling} and {Its} {Applications}: {Monographs} on {Statistics} and {Applied} {Probability} 66}.
\newblock Routledge, New York, May 2018.
\newblock ISBN 978-0-203-74872-5.
\newblock \doi{10.1201/9780203748725}.

\bibitem[Fu et~al.(2022)Fu, Dao, Saab, Thomas, Rudra, and Re]{Fu2022HungryHungryHippos}
D.~Y. Fu, T.~Dao, K.~K. Saab, A.~W. Thomas, A.~Rudra, and C.~Re.
\newblock Hungry {Hungry} {Hippos}: {Towards} {Language} {Modeling} with {State} {Space} {Models}.
\newblock In \emph{The {Eleventh} {International} {Conference} on {Learning} {Representations}}, Sept. 2022.
\newblock URL \url{https://openreview.net/forum?id=COZDy0WYGg}.

\bibitem[Garg et~al.(2022)Garg, Tsipras, Liang, and Valiant]{Garg2022WhatCanTransformers}
S.~Garg, D.~Tsipras, P.~Liang, and G.~Valiant.
\newblock What can transformers learn in-context? {A} case study of simple function classes.
\newblock In A.~H. Oh, A.~Agarwal, D.~Belgrave, and K.~Cho, editors, \emph{Advances in neural information processing systems}, 2022.
\newblock URL \url{https://openreview.net/forum?id=flNZJ2eOet}.

\bibitem[Garnelo and Czarnecki(2023)]{Garnelo2023ExploringSpaceKeyValueQuery}
M.~Garnelo and W.~M. Czarnecki.
\newblock Exploring the {Space} of {Key}-{Value}-{Query} {Models} with {Intention}, May 2023.
\newblock URL \url{http://arxiv.org/abs/2305.10203}.
\newblock arXiv:2305.10203.

\bibitem[Gehring et~al.(2017)Gehring, Auli, Grangier, Yarats, and Dauphin]{Gehring2017ConvolutionalSequenceSequence}
J.~Gehring, M.~Auli, D.~Grangier, D.~Yarats, and Y.~N. Dauphin.
\newblock Convolutional {Sequence} to {Sequence} {Learning}.
\newblock In \emph{Proceedings of the 34th {International} {Conference} on {Machine} {Learning}}, pages 1243--1252. PMLR, July 2017.
\newblock URL \url{https://proceedings.mlr.press/v70/gehring17a.html}.
\newblock ISSN: 2640-3498.

\bibitem[Goel and Bartlett(2024)]{Goel2024CanTransformerRepresent}
G.~Goel and P.~Bartlett.
\newblock Can a transformer represent a {Kalman} filter?
\newblock In A.~Abate, M.~Cannon, K.~Margellos, and A.~Papachristodoulou, editors, \emph{Proceedings of the 6th annual learning for dynamics \&amp; control conference}, volume 242 of \emph{Proceedings of machine learning research}, pages 1502--1512. PMLR, July 2024.
\newblock URL \url{https://proceedings.mlr.press/v242/goel24a.html}.

\bibitem[Greff et~al.(2017)Greff, Srivastava, Koutník, Steunebrink, and Schmidhuber]{Greff2017LSTMSearchSpace}
K.~Greff, R.~K. Srivastava, J.~Koutník, B.~R. Steunebrink, and J.~Schmidhuber.
\newblock {LSTM}: {A} {Search} {Space} {Odyssey}.
\newblock \emph{IEEE Transactions on Neural Networks and Learning Systems}, 28\penalty0 (10):\penalty0 2222--2232, Oct. 2017.
\newblock ISSN 2162-237X, 2162-2388.
\newblock \doi{10.1109/TNNLS.2016.2582924}.
\newblock URL \url{http://arxiv.org/abs/1503.04069}.
\newblock arXiv:1503.04069 [cs].

\bibitem[Gruver et~al.(2023)Gruver, Finzi, Qiu, and Wilson]{Gruver2023LargeLanguageModels}
N.~Gruver, M.~Finzi, S.~Qiu, and A.~G. Wilson.
\newblock Large {Language} {Models} {Are} {Zero}-{Shot} {Time} {Series} {Forecasters}.
\newblock \emph{Advances in Neural Information Processing Systems}, 36:\penalty0 19622--19635, Dec. 2023.
\newblock URL \url{https://proceedings.neurips.cc/paper_files/paper/2023/hash/3eb7ca52e8207697361b2c0fb3926511-Abstract-Conference.html}.

\bibitem[Gu and Dao(2024)]{Gu2024MambaLinearTimeSequence}
A.~Gu and T.~Dao.
\newblock Mamba: {Linear}-{Time} {Sequence} {Modeling} with {Selective} {State} {Spaces}.
\newblock In \emph{First {Conference} on {Language} {Modeling}}, Aug. 2024.
\newblock URL \url{https://openreview.net/forum?id=tEYskw1VY2#discussion}.

\bibitem[Gu et~al.(2020)Gu, Dao, Ermon, Rudra, and Ré]{Gu2020HiPPORecurrentMemory}
A.~Gu, T.~Dao, S.~Ermon, A.~Rudra, and C.~Ré.
\newblock {HiPPO}: {Recurrent} {Memory} with {Optimal} {Polynomial} {Projections}.
\newblock In \emph{Advances in {Neural} {Information} {Processing} {Systems}}, volume~33, pages 1474--1487. Curran Associates, Inc., 2020.
\newblock URL \url{https://proceedings.neurips.cc/paper/2020/hash/102f0bb6efb3a6128a3c750dd16729be-Abstract.html}.

\bibitem[Gu et~al.(2021)Gu, Goel, and Re]{Gu2021EfficientlyModelingLong}
A.~Gu, K.~Goel, and C.~Re.
\newblock Efficiently {Modeling} {Long} {Sequences} with {Structured} {State} {Spaces}.
\newblock In \emph{International {Conference} on {Learning} {Representations}}, Oct. 2021.
\newblock URL \url{https://openreview.net/forum?id=uYLFoz1vlAC}.

\bibitem[Gu et~al.(2022)Gu, Gupta, Goel, and Ré]{Gu2022ParameterizationInitializationDiagonal}
A.~Gu, A.~Gupta, K.~Goel, and C.~Ré.
\newblock On the parameterization and initialization of diagonal state space models.
\newblock In \emph{Proceedings of the 36th {International} {Conference} on {Neural} {Information} {Processing} {Systems}}, {NIPS} '22, pages 35971--35983, Red Hook, NY, USA, 2022. Curran Associates Inc.
\newblock ISBN 978-1-7138-7108-8.

\bibitem[Han et~al.(2024)Han, Wang, Xia, Han, Pu, Ge, Song, Song, Zheng, and Huang]{Han2024DemystifyMambaVision}
D.~Han, Z.~Wang, Z.~Xia, Y.~Han, Y.~Pu, C.~Ge, J.~Song, S.~Song, B.~Zheng, and G.~Huang.
\newblock Demystify {Mamba} in {Vision}: {A} {Linear} {Attention} {Perspective}, May 2024.
\newblock URL \url{https://arxiv.org/abs/2405.16605v1}.

\bibitem[Haykin(2014)]{Haykin2014AdaptiveFilterTheory}
S.~S. Haykin.
\newblock \emph{Adaptive {Filter} {Theory}}.
\newblock Pearson, 2014.
\newblock ISBN 978-0-13-267145-3.
\newblock Google-Books-ID: J4GRKQEACAAJ.

\bibitem[Hinton and Anderson(1989)]{Hinton1989ParallelModelsAssociative}
G.~E. Hinton and J.~A. Anderson.
\newblock \emph{Parallel {Models} of {Associative} {Memory}}.
\newblock Psychology Press, 1989.

\bibitem[Hochreiter and Schmidhuber(1997)]{Hochreiter1997LongShortTermMemory}
S.~Hochreiter and J.~Schmidhuber.
\newblock Long {Short}-{Term} {Memory}.
\newblock \emph{Neural Computation}, 9\penalty0 (8):\penalty0 1735--1780, Nov. 1997.
\newblock ISSN 0899-7667.
\newblock \doi{10.1162/neco.1997.9.8.1735}.
\newblock URL \url{https://ieeexplore.ieee.org/abstract/document/6795963}.
\newblock Conference Name: Neural Computation.

\bibitem[Hoffmann et~al.(2022)Hoffmann, Borgeaud, Mensch, Buchatskaya, Cai, Rutherford, Casas, Hendricks, Welbl, Clark, Hennigan, Noland, Millican, Driessche, Damoc, Guy, Osindero, Simonyan, Elsen, Rae, Vinyals, and Sifre]{Hoffmann2022TrainingComputeOptimalLarge}
J.~Hoffmann, S.~Borgeaud, A.~Mensch, E.~Buchatskaya, T.~Cai, E.~Rutherford, D.~d.~L. Casas, L.~A. Hendricks, J.~Welbl, A.~Clark, T.~Hennigan, E.~Noland, K.~Millican, G.~v.~d. Driessche, B.~Damoc, A.~Guy, S.~Osindero, K.~Simonyan, E.~Elsen, J.~W. Rae, O.~Vinyals, and L.~Sifre.
\newblock Training {Compute}-{Optimal} {Large} {Language} {Models}, Mar. 2022.
\newblock URL \url{http://arxiv.org/abs/2203.15556}.
\newblock arXiv:2203.15556 [cs].

\bibitem[Hopfield(1982)]{Hopfield1982NeuralNetworksPhysical}
J.~J. Hopfield.
\newblock Neural networks and physical systems with emergent collective computational abilities.
\newblock \emph{Proceedings of the National Academy of Sciences}, 79\penalty0 (8):\penalty0 2554--2558, Apr. 1982.
\newblock \doi{10.1073/pnas.79.8.2554}.
\newblock URL \url{https://www.pnas.org/doi/abs/10.1073/pnas.79.8.2554}.
\newblock Publisher: Proceedings of the National Academy of Sciences.

\bibitem[Hua et~al.(2022)Hua, Dai, Liu, and Le]{Hua2022TransformerQualityLinear}
W.~Hua, Z.~Dai, H.~Liu, and Q.~Le.
\newblock Transformer {Quality} in {Linear} {Time}.
\newblock In \emph{Proceedings of the 39th {International} {Conference} on {Machine} {Learning}}, pages 9099--9117. PMLR, June 2022.
\newblock URL \url{https://proceedings.mlr.press/v162/hua22a.html}.
\newblock ISSN: 2640-3498.

\bibitem[Johnstone et~al.(1982)Johnstone, Johnson, Bitmead, and O.~Anderson]{Johnstone1982ExponentialConvergenceRecursive}
R.~M. Johnstone, C.~R. Johnson, R.~R. Bitmead, and B.~D. O.~Anderson.
\newblock Exponential convergence of recursive least squares with exponential forgetting factor.
\newblock In \emph{1982 21st {IEEE} {Conference} on {Decision} and {Control}}, pages 994--997, Dec. 1982.
\newblock \doi{10.1109/CDC.1982.268295}.
\newblock URL \url{https://ieeexplore.ieee.org/document/4047398}.

\bibitem[Jumper et~al.(2021)Jumper, Evans, Pritzel, Green, Figurnov, Ronneberger, Tunyasuvunakool, Bates, Žídek, Potapenko, Bridgland, Meyer, Kohl, Ballard, Cowie, Romera-Paredes, Nikolov, Jain, Adler, Back, Petersen, Reiman, Clancy, Zielinski, Steinegger, Pacholska, Berghammer, Bodenstein, Silver, Vinyals, Senior, Kavukcuoglu, Kohli, and Hassabis]{Jumper2021HighlyAccurateProtein}
J.~Jumper, R.~Evans, A.~Pritzel, T.~Green, M.~Figurnov, O.~Ronneberger, K.~Tunyasuvunakool, R.~Bates, A.~Žídek, A.~Potapenko, A.~Bridgland, C.~Meyer, S.~A.~A. Kohl, A.~J. Ballard, A.~Cowie, B.~Romera-Paredes, S.~Nikolov, R.~Jain, J.~Adler, T.~Back, S.~Petersen, D.~Reiman, E.~Clancy, M.~Zielinski, M.~Steinegger, M.~Pacholska, T.~Berghammer, S.~Bodenstein, D.~Silver, O.~Vinyals, A.~W. Senior, K.~Kavukcuoglu, P.~Kohli, and D.~Hassabis.
\newblock Highly accurate protein structure prediction with {AlphaFold}.
\newblock \emph{Nature}, 596\penalty0 (7873):\penalty0 583--589, Aug. 2021.
\newblock ISSN 1476-4687.
\newblock \doi{10.1038/s41586-021-03819-2}.
\newblock URL \url{https://www.nature.com/articles/s41586-021-03819-2}.
\newblock Publisher: Nature Publishing Group.

\bibitem[Kaczmarz(1937)]{Kaczmarz1937AngenaherteAuflosungSystemen}
S.~Kaczmarz.
\newblock Angenaherte auflosung von systemen linearer glei-chungen.
\newblock \emph{Bull. Int. Acad. Pol. Sic. Let., Cl. Sci. Math. Nat.}, pages 355--357, 1937.
\newblock URL \url{https://cir.nii.ac.jp/crid/1570009749812709888}.

\bibitem[Kaplan et~al.(2020)Kaplan, McCandlish, Henighan, Brown, Chess, Child, Gray, Radford, Wu, and Amodei]{Kaplan2020ScalingLawsNeural}
J.~Kaplan, S.~McCandlish, T.~Henighan, T.~B. Brown, B.~Chess, R.~Child, S.~Gray, A.~Radford, J.~Wu, and D.~Amodei.
\newblock Scaling {Laws} for {Neural} {Language} {Models}, Jan. 2020.
\newblock URL \url{http://arxiv.org/abs/2001.08361}.
\newblock arXiv:2001.08361 [cs].

\bibitem[Kasai et~al.(2021)Kasai, Peng, Zhang, Yogatama, Ilharco, Pappas, Mao, Chen, and Smith]{Kasai2021FinetuningPretrainedTransformers}
J.~Kasai, H.~Peng, Y.~Zhang, D.~Yogatama, G.~Ilharco, N.~Pappas, Y.~Mao, W.~Chen, and N.~A. Smith.
\newblock Finetuning {Pretrained} {Transformers} into {RNNs}.
\newblock In M.-F. Moens, X.~Huang, L.~Specia, and S.~W.-t. Yih, editors, \emph{Proceedings of the 2021 {Conference} on {Empirical} {Methods} in {Natural} {Language} {Processing}}, pages 10630--10643, Online and Punta Cana, Dominican Republic, Nov. 2021. Association for Computational Linguistics.
\newblock \doi{10.18653/v1/2021.emnlp-main.830}.
\newblock URL \url{https://aclanthology.org/2021.emnlp-main.830}.

\bibitem[Katharopoulos et~al.(2020)Katharopoulos, Vyas, Pappas, and Fleuret]{Katharopoulos2020TransformersAreRNNs}
A.~Katharopoulos, A.~Vyas, N.~Pappas, and F.~Fleuret.
\newblock Transformers are {RNNs}: {Fast} {Autoregressive} {Transformers} with {Linear} {Attention}.
\newblock In \emph{Proceedings of the 37th {International} {Conference} on {Machine} {Learning}}, pages 5156--5165. PMLR, Nov. 2020.
\newblock URL \url{https://proceedings.mlr.press/v119/katharopoulos20a.html}.
\newblock ISSN: 2640-3498.

\bibitem[Katsch(2024)]{Katsch2024GateLoopFullyDataControlled}
T.~Katsch.
\newblock {GateLoop}: {Fully} {Data}-{Controlled} {Linear} {Recurrence} for {Sequence} {Modeling}, Jan. 2024.
\newblock URL \url{http://arxiv.org/abs/2311.01927}.
\newblock arXiv:2311.01927 [cs].

\bibitem[Kingma and Ba(2015)]{Kingma2015AdamMethodStochastic}
D.~P. Kingma and J.~Ba.
\newblock Adam: {A} {Method} for {Stochastic} {Optimization}.
\newblock 2015.
\newblock URL \url{https://arxiv.org/abs/1412.6980}.

\bibitem[Kohonen(1972)]{Kohonen1972CorrelationMatrixMemories}
T.~Kohonen.
\newblock Correlation {Matrix} {Memories}.
\newblock \emph{IEEE Transactions on Computers}, C-21\penalty0 (4):\penalty0 353--359, Apr. 1972.
\newblock ISSN 1557-9956.
\newblock \doi{10.1109/TC.1972.5008975}.
\newblock URL \url{https://ieeexplore.ieee.org/abstract/document/5008975}.
\newblock Conference Name: IEEE Transactions on Computers.

\bibitem[Kohonen(1989)]{Kohonen1989SelfOrganizationAssociativeMemory}
T.~Kohonen.
\newblock \emph{Self-{Organization} and {Associative} {Memory}}, volume~8 of \emph{Springer {Series} in {Information} {Sciences}}.
\newblock Springer, Berlin, Heidelberg, 1989.
\newblock ISBN 978-3-540-51387-2 978-3-642-88163-3.
\newblock \doi{10.1007/978-3-642-88163-3}.
\newblock URL \url{http://link.springer.com/10.1007/978-3-642-88163-3}.

\bibitem[Krotov and Hopfield(2021)]{Krotov2021LargeAssociativeMemory}
D.~Krotov and J.~Hopfield.
\newblock Large {Associative} {Memory} {Problem} in {Neurobiology} and {Machine} {Learning}, Apr. 2021.
\newblock URL \url{http://arxiv.org/abs/2008.06996}.
\newblock arXiv:2008.06996.

\bibitem[Krotov and Hopfield(2016)]{Krotov2016DenseAssociativeMemory}
D.~Krotov and J.~J. Hopfield.
\newblock Dense {Associative} {Memory} for {Pattern} {Recognition}.
\newblock In \emph{Advances in {Neural} {Information} {Processing} {Systems}}, volume~29. Curran Associates, Inc., 2016.
\newblock URL \url{https://proceedings.neurips.cc/paper_files/paper/2016/hash/eaae339c4d89fc102edd9dbdb6a28915-Abstract.html}.

\bibitem[Liu et~al.(2024)Liu, Wang, Wu, Feng, Stone, and Liu]{Liu2024LonghornStateSpace}
B.~Liu, R.~Wang, L.~Wu, Y.~Feng, P.~Stone, and Q.~Liu.
\newblock Longhorn: {State} {Space} {Models} are {Amortized} {Online} {Learners}, Oct. 2024.
\newblock URL \url{http://arxiv.org/abs/2407.14207}.
\newblock arXiv:2407.14207.

\bibitem[Loshchilov and Hutter(2019)]{Loshchilov2019DecoupledWeightDecay}
I.~Loshchilov and F.~Hutter.
\newblock Decoupled weight decay regularization.
\newblock In \emph{International conference on learning representations}, 2019.
\newblock URL \url{https://openreview.net/forum?id=Bkg6RiCqY7}.

\bibitem[Lu et~al.(2021)Lu, Yao, Zhang, Zhu, Xu, Gao, XU, Xiang, and Zhang]{Lu2021SOFTSoftmaxfreeTransformer}
J.~Lu, J.~Yao, J.~Zhang, X.~Zhu, H.~Xu, W.~Gao, C.~XU, T.~Xiang, and L.~Zhang.
\newblock {SOFT}: {Softmax}-free {Transformer} with {Linear} {Complexity}.
\newblock In \emph{Advances in {Neural} {Information} {Processing} {Systems}}, volume~34, pages 21297--21309. Curran Associates, Inc., 2021.
\newblock URL \url{https://proceedings.neurips.cc/paper/2021/hash/b1d10e7bafa4421218a51b1e1f1b0ba2-Abstract.html}.

\bibitem[Mehta et~al.(2023)Mehta, Gupta, Cutkosky, and Neyshabur]{Mehta2023LongRangeLanguage}
H.~Mehta, A.~Gupta, A.~Cutkosky, and B.~Neyshabur.
\newblock Long range language modeling via gated state spaces.
\newblock In \emph{The eleventh international conference on learning representations}, 2023.
\newblock URL \url{https://openreview.net/forum?id=5MkYIYCbva}.

\bibitem[Millidge et~al.(2022)Millidge, Salvatori, Song, Lukasiewicz, and Bogacz]{Millidge2022UniversalHopfieldNetworks}
B.~Millidge, T.~Salvatori, Y.~Song, T.~Lukasiewicz, and R.~Bogacz.
\newblock Universal {Hopfield} {Networks}: {A} {General} {Framework} for {Single}-{Shot} {Associative} {Memory} {Models}.
\newblock In \emph{Proceedings of the 39th {International} {Conference} on {Machine} {Learning}}, pages 15561--15583. PMLR, June 2022.
\newblock URL \url{https://proceedings.mlr.press/v162/millidge22a.html}.
\newblock ISSN: 2640-3498.

\bibitem[Nguyen et~al.(2024)Nguyen, Poli, Durrant, Kang, Katrekar, Li, Bartie, Thomas, King, Brixi, Sullivan, Ng, Lewis, Lou, Ermon, Baccus, Hernandez-Boussard, Ré, Hsu, and Hie]{Nguyen2024SequenceModelingDesign}
E.~Nguyen, M.~Poli, M.~G. Durrant, B.~Kang, D.~Katrekar, D.~B. Li, L.~J. Bartie, A.~W. Thomas, S.~H. King, G.~Brixi, J.~Sullivan, M.~Y. Ng, A.~Lewis, A.~Lou, S.~Ermon, S.~A. Baccus, T.~Hernandez-Boussard, C.~Ré, P.~D. Hsu, and B.~L. Hie.
\newblock Sequence modeling and design from molecular to genome scale with {Evo}.
\newblock \emph{Science}, 386\penalty0 (6723):\penalty0 eado9336, Nov. 2024.
\newblock \doi{10.1126/science.ado9336}.
\newblock URL \url{https://www.science.org/doi/10.1126/science.ado9336}.
\newblock Publisher: American Association for the Advancement of Science.

\bibitem[Nichani et~al.(2024)Nichani, Lee, and Bietti]{Nichani2024UnderstandingFactualRecall}
E.~Nichani, J.~D. Lee, and A.~Bietti.
\newblock Understanding {Factual} {Recall} in {Transformers} via {Associative} {Memories}, Dec. 2024.
\newblock URL \url{http://arxiv.org/abs/2412.06538}.
\newblock arXiv:2412.06538 [cs].

\bibitem[Olsson et~al.(2022)Olsson, Elhage, Nanda, Joseph, DasSarma, Henighan, Mann, Askell, Bai, Chen, Conerly, Drain, Ganguli, Hatfield-Dodds, Hernandez, Johnston, Jones, Kernion, Lovitt, Ndousse, Amodei, Brown, Clark, Kaplan, McCandlish, and Olah]{Olsson2022ContextLearningInduction}
C.~Olsson, N.~Elhage, N.~Nanda, N.~Joseph, N.~DasSarma, T.~Henighan, B.~Mann, A.~Askell, Y.~Bai, A.~Chen, T.~Conerly, D.~Drain, D.~Ganguli, Z.~Hatfield-Dodds, D.~Hernandez, S.~Johnston, A.~Jones, J.~Kernion, L.~Lovitt, K.~Ndousse, D.~Amodei, T.~Brown, J.~Clark, J.~Kaplan, S.~McCandlish, and C.~Olah.
\newblock In-context {Learning} and {Induction} {Heads}, Sept. 2022.
\newblock URL \url{http://arxiv.org/abs/2209.11895}.
\newblock arXiv:2209.11895 [cs].

\bibitem[OpenAI(2024)]{OpenAI2024OpenAIO1System}
OpenAI.
\newblock {OpenAI} o1 {System} {Card}, 2024.
\newblock URL \url{https://openai.com/index/openai-o1-system-card/}.

\bibitem[Orvieto et~al.(2023)Orvieto, Smith, Gu, Fernando, Gulcehre, Pascanu, and De]{Orvieto2023ResurrectingRecurrentNeural}
A.~Orvieto, S.~L. Smith, A.~Gu, A.~Fernando, C.~Gulcehre, R.~Pascanu, and S.~De.
\newblock Resurrecting {Recurrent} {Neural} {Networks} for {Long} {Sequences}.
\newblock In \emph{Proceedings of the 40th {International} {Conference} on {Machine} {Learning}}, pages 26670--26698. PMLR, July 2023.
\newblock URL \url{https://proceedings.mlr.press/v202/orvieto23a.html}.
\newblock ISSN: 2640-3498.

\bibitem[Oswald et~al.(2023)Oswald, Niklasson, Randazzo, Sacramento, Mordvintsev, Zhmoginov, and Vladymyrov]{Oswald2023TransformersLearnInContext}
J.~V. Oswald, E.~Niklasson, E.~Randazzo, J.~Sacramento, A.~Mordvintsev, A.~Zhmoginov, and M.~Vladymyrov.
\newblock Transformers {Learn} {In}-{Context} by {Gradient} {Descent}.
\newblock In \emph{Proceedings of the 40th {International} {Conference} on {Machine} {Learning}}, pages 35151--35174. PMLR, July 2023.
\newblock URL \url{https://proceedings.mlr.press/v202/von-oswald23a.html}.
\newblock ISSN: 2640-3498.

\bibitem[Peng et~al.(2024)Peng, Goldstein, Anthony, Albalak, Alcaide, Biderman, Cheah, Du, Ferdinan, Hou, Kazienko, GV, Kocoń, Koptyra, Krishna, Jr, Lin, Muennighoff, Obeid, Saito, Song, Tu, Wirawan, Woźniak, Zhang, Zhao, Zhao, Zhou, Zhu, and Zhu]{Peng2024EagleFinchRWKV}
B.~Peng, D.~Goldstein, Q.~Anthony, A.~Albalak, E.~Alcaide, S.~Biderman, E.~Cheah, X.~Du, T.~Ferdinan, H.~Hou, P.~Kazienko, K.~K. GV, J.~Kocoń, B.~Koptyra, S.~Krishna, R.~M. Jr, J.~Lin, N.~Muennighoff, F.~Obeid, A.~Saito, G.~Song, H.~Tu, C.~Wirawan, S.~Woźniak, R.~Zhang, B.~Zhao, Q.~Zhao, P.~Zhou, J.~Zhu, and R.-J. Zhu.
\newblock Eagle and {Finch}: {RWKV} with {Matrix}-{Valued} {States} and {Dynamic} {Recurrence}, Sept. 2024.
\newblock URL \url{http://arxiv.org/abs/2404.05892}.
\newblock arXiv:2404.05892.

\bibitem[Peng et~al.(2020)Peng, Pappas, Yogatama, Schwartz, Smith, and Kong]{Peng2020RandomFeatureAttention}
H.~Peng, N.~Pappas, D.~Yogatama, R.~Schwartz, N.~Smith, and L.~Kong.
\newblock Random {Feature} {Attention}.
\newblock In \emph{International {Conference} on {Learning} {Representations}}, Oct. 2020.
\newblock URL \url{https://openreview.net/forum?id=QtTKTdVrFBB}.

\bibitem[Poggio(1975)]{Poggio1975OptimalNonlinearAssociative}
T.~Poggio.
\newblock On optimal nonlinear associative recall.
\newblock \emph{Biological Cybernetics}, 19\penalty0 (4):\penalty0 201--209, Sept. 1975.
\newblock ISSN 1432-0770.
\newblock \doi{10.1007/BF02281970}.
\newblock URL \url{https://doi.org/10.1007/BF02281970}.

\bibitem[Poli et~al.(2023)Poli, Massaroli, Nguyen, Fu, Dao, Baccus, Bengio, Ermon, and Ré]{Poli2023HyenaHierarchyLarger}
M.~Poli, S.~Massaroli, E.~Nguyen, D.~Y. Fu, T.~Dao, S.~Baccus, Y.~Bengio, S.~Ermon, and C.~Ré.
\newblock Hyena hierarchy: towards larger convolutional language models.
\newblock In \emph{Proceedings of the 40th {International} {Conference} on {Machine} {Learning}}, volume 202 of \emph{{ICML}'23}, pages 28043--28078, Honolulu, Hawaii, USA, July 2023. JMLR.org.

\bibitem[Qin et~al.(2021)Qin, Sun, Deng, Li, Wei, Lv, Yan, Kong, and Zhong]{Qin2021CosFormerRethinkingSoftmax}
Z.~Qin, W.~Sun, H.~Deng, D.~Li, Y.~Wei, B.~Lv, J.~Yan, L.~Kong, and Y.~Zhong.
\newblock {cosFormer}: {Rethinking} {Softmax} {In} {Attention}.
\newblock In \emph{International {Conference} on {Learning} {Representations}}, Oct. 2021.
\newblock URL \url{https://openreview.net/forum?id=Bl8CQrx2Up4}.

\bibitem[Qin et~al.(2022)Qin, Han, Sun, Li, Kong, Barnes, and Zhong]{Qin2022DevilLinearTransformer}
Z.~Qin, X.~Han, W.~Sun, D.~Li, L.~Kong, N.~Barnes, and Y.~Zhong.
\newblock The {Devil} in {Linear} {Transformer}.
\newblock In Y.~Goldberg, Z.~Kozareva, and Y.~Zhang, editors, \emph{Proceedings of the 2022 {Conference} on {Empirical} {Methods} in {Natural} {Language} {Processing}}, pages 7025--7041, Abu Dhabi, United Arab Emirates, Dec. 2022. Association for Computational Linguistics.
\newblock \doi{10.18653/v1/2022.emnlp-main.473}.
\newblock URL \url{https://aclanthology.org/2022.emnlp-main.473}.

\bibitem[Qin et~al.(2023)Qin, Han, Sun, He, Li, Li, Dai, Kong, and Zhong]{Qin2023ToeplitzNeuralNetwork}
Z.~Qin, X.~Han, W.~Sun, B.~He, D.~Li, D.~Li, Y.~Dai, L.~Kong, and Y.~Zhong.
\newblock Toeplitz neural network for sequence modeling.
\newblock In \emph{The eleventh international conference on learning representations}, 2023.
\newblock URL \url{https://openreview.net/forum?id=IxmWsm4xrua}.

\bibitem[Qin et~al.(2024)Qin, Yang, Sun, Shen, Li, Sun, and Zhong]{Qin2024HGRN2GatedLinear}
Z.~Qin, S.~Yang, W.~Sun, X.~Shen, D.~Li, W.~Sun, and Y.~Zhong.
\newblock {HGRN2}: {Gated} {Linear} {RNNs} with {State} {Expansion}.
\newblock In \emph{First {Conference} on {Language} {Modeling}}, Aug. 2024.
\newblock URL \url{https://openreview.net/forum?id=y6SqbJfCSk#discussion}.

\bibitem[Ramsauer et~al.(2021)Ramsauer, Schäfl, Lehner, Seidl, Widrich, Gruber, Holzleitner, Adler, Kreil, Kopp, Klambauer, Brandstetter, and Hochreiter]{Ramsauer2020HopfieldNetworksAll}
H.~Ramsauer, B.~Schäfl, J.~Lehner, P.~Seidl, M.~Widrich, L.~Gruber, M.~Holzleitner, T.~Adler, D.~Kreil, M.~K. Kopp, G.~Klambauer, J.~Brandstetter, and S.~Hochreiter.
\newblock Hopfield {Networks} is {All} {You} {Need}.
\newblock In \emph{International {Conference} on {Learning} {Representations}}, 2021.
\newblock URL \url{https://openreview.net/forum?id=tL89RnzIiCd}.

\bibitem[Ren et~al.(2023)Ren, Liu, Wang, Xu, Zhu, and Zhai]{Ren2023SparseModularActivation}
L.~Ren, Y.~Liu, S.~Wang, Y.~Xu, C.~Zhu, and C.~Zhai.
\newblock Sparse modular activation for efficient sequence modeling.
\newblock In \emph{Thirty-seventh conference on neural information processing systems}, 2023.
\newblock URL \url{https://openreview.net/forum?id=TfbzX6I14i}.

\bibitem[Ren et~al.(2024)Ren, Li, and Liu]{Ren2024CanMambaAlways}
R.~Ren, Z.~Li, and Y.~Liu.
\newblock Can {Mamba} {Always} {Enjoy} the "{Free} {Lunch}"?, Oct. 2024.
\newblock URL \url{http://arxiv.org/abs/2410.03810}.
\newblock arXiv:2410.03810 [cs].

\bibitem[Rush(2024)]{SashaRush2024ThereAre4}
S.~Rush.
\newblock There are like 4 more linear {RNN} papers out today, Apr. 2024.
\newblock URL \url{https://x.com/srush_nlp/status/1780231813820002771}.

\bibitem[Salinas et~al.(2020)Salinas, Flunkert, Gasthaus, and Januschowski]{Salinas2020DeepARProbabilisticForecasting}
D.~Salinas, V.~Flunkert, J.~Gasthaus, and T.~Januschowski.
\newblock {DeepAR}: {Probabilistic} forecasting with autoregressive recurrent networks.
\newblock \emph{International Journal of Forecasting}, 36\penalty0 (3):\penalty0 1181--1191, July 2020.
\newblock ISSN 0169-2070.
\newblock \doi{10.1016/j.ijforecast.2019.07.001}.
\newblock URL \url{https://www.sciencedirect.com/science/article/pii/S0169207019301888}.

\bibitem[Schlag et~al.(2021)Schlag, Irie, and Schmidhuber]{Schlag2021LinearTransformersAre}
I.~Schlag, K.~Irie, and J.~Schmidhuber.
\newblock Linear {Transformers} {Are} {Secretly} {Fast} {Weight} {Programmers}.
\newblock In \emph{Proceedings of the 38th {International} {Conference} on {Machine} {Learning}}, pages 9355--9366. PMLR, July 2021.
\newblock URL \url{https://proceedings.mlr.press/v139/schlag21a.html}.
\newblock ISSN: 2640-3498.

\bibitem[Schmidhuber(1992)]{Schmidhuber1992LearningControlFastWeight}
J.~Schmidhuber.
\newblock Learning to {Control} {Fast}-{Weight} {Memories}: {An} {Alternative} to {Dynamic} {Recurrent} {Networks}.
\newblock \emph{Neural Computation}, 4\penalty0 (1):\penalty0 131--139, Jan. 1992.
\newblock ISSN 0899-7667.
\newblock \doi{10.1162/neco.1992.4.1.131}.
\newblock URL \url{https://doi.org/10.1162/neco.1992.4.1.131}.

\bibitem[Schölkopf and Smola(2001)]{Scholkopf2001LearningKernelsSupport}
B.~Schölkopf and A.~J. Smola.
\newblock \emph{Learning with {Kernels}: {Support} {Vector} {Machines}, {Regularization}, {Optimization}, and {Beyond}}.
\newblock The MIT Press, Dec. 2001.
\newblock ISBN 978-0-262-25693-3.
\newblock \doi{10.7551/mitpress/4175.001.0001}.
\newblock URL \url{https://direct.mit.edu/books/monograph/1821/Learning-with-KernelsSupport-Vector-Machines}.

\bibitem[Shazeer(2019)]{Shazeer2019FastTransformerDecoding}
N.~Shazeer.
\newblock Fast {Transformer} {Decoding}: {One} {Write}-{Head} is {All} {You} {Need}, Nov. 2019.
\newblock URL \url{http://arxiv.org/abs/1911.02150}.
\newblock arXiv:1911.02150 [cs].

\bibitem[Shi et~al.(2023)Shi, Wang, and Fox]{Shi2023SequenceModelingMultiresolutiona}
J.~Shi, K.~A. Wang, and E.~B. Fox.
\newblock Sequence modeling with multiresolution convolutional memory.
\newblock In \emph{Proceedings of the 40th {International} {Conference} on {Machine} {Learning}}, volume 202 of \emph{{ICML}'23}, pages 31312--31327, Honolulu, Hawaii, USA, July 2023. JMLR.org.

\bibitem[Siems et~al.(2025)Siems, Carstensen, Zela, Hutter, Pontil, and Grazzi]{Siems2025DeltaProductIncreasingExpressivity}
J.~Siems, T.~Carstensen, A.~Zela, F.~Hutter, M.~Pontil, and R.~Grazzi.
\newblock {DeltaProduct}: {Increasing} the {Expressivity} of {DeltaNet} {Through} {Products} of {Householders}, Feb. 2025.
\newblock URL \url{http://arxiv.org/abs/2502.10297}.
\newblock arXiv:2502.10297 [cs].

\bibitem[Smith et~al.(2023)Smith, Warrington, and Linderman]{Smith2023SimplifiedStateSpace}
J.~T. Smith, A.~Warrington, and S.~Linderman.
\newblock Simplified state space layers for sequence modeling.
\newblock In \emph{The eleventh international conference on learning representations}, 2023.
\newblock URL \url{https://openreview.net/forum?id=Ai8Hw3AXqks}.

\bibitem[Sukhbaatar et~al.(2019)Sukhbaatar, Grave, Lample, Jegou, and Joulin]{Sukhbaatar2019AugmentingSelfattentionPersistent}
S.~Sukhbaatar, E.~Grave, G.~Lample, H.~Jegou, and A.~Joulin.
\newblock Augmenting {Self}-attention with {Persistent} {Memory}, July 2019.
\newblock URL \url{http://arxiv.org/abs/1907.01470}.
\newblock arXiv:1907.01470.

\bibitem[Sun et~al.(2020)Sun, Wang, Liu, Miller, Efros, and Hardt]{Sun2020TestTimeTrainingSelfSupervision}
Y.~Sun, X.~Wang, Z.~Liu, J.~Miller, A.~Efros, and M.~Hardt.
\newblock Test-{Time} {Training} with {Self}-{Supervision} for {Generalization} under {Distribution} {Shifts}.
\newblock In \emph{Proceedings of the 37th {International} {Conference} on {Machine} {Learning}}, pages 9229--9248. PMLR, Nov. 2020.
\newblock URL \url{https://proceedings.mlr.press/v119/sun20b.html}.
\newblock ISSN: 2640-3498.

\bibitem[Sun et~al.(2023)Sun, Dong, Huang, Ma, Xia, Xue, Wang, and Wei]{Sun2023RetentiveNetworkSuccessor}
Y.~Sun, L.~Dong, S.~Huang, S.~Ma, Y.~Xia, J.~Xue, J.~Wang, and F.~Wei.
\newblock Retentive {Network}: {A} {Successor} to {Transformer} for {Large} {Language} {Models}, Aug. 2023.
\newblock URL \url{http://arxiv.org/abs/2307.08621}.
\newblock arXiv:2307.08621 [cs].

\bibitem[Sun et~al.(2024)Sun, Li, Dalal, Xu, Vikram, Zhang, Dubois, Chen, Wang, Koyejo, Hashimoto, and Guestrin]{Sun2024LearningLearnTest}
Y.~Sun, X.~Li, K.~Dalal, J.~Xu, A.~Vikram, G.~Zhang, Y.~Dubois, X.~Chen, X.~Wang, S.~Koyejo, T.~Hashimoto, and C.~Guestrin.
\newblock Learning to ({Learn} at {Test} {Time}): {RNNs} with {Expressive} {Hidden} {States}, Aug. 2024.
\newblock URL \url{http://arxiv.org/abs/2407.04620}.
\newblock arXiv:2407.04620.

\bibitem[Sutskever et~al.(2014)Sutskever, Vinyals, and Le]{Sutskever2014SequenceSequenceLearning}
I.~Sutskever, O.~Vinyals, and Q.~V. Le.
\newblock Sequence to {Sequence} {Learning} with {Neural} {Networks}.
\newblock In \emph{Advances in {Neural} {Information} {Processing} {Systems}}, volume~27. Curran Associates, Inc., 2014.
\newblock URL \url{https://proceedings.neurips.cc/paper/2014/hash/a14ac55a4f27472c5d894ec1c3c743d2-Abstract.html}.

\bibitem[Sutton(2019)]{Sutton2019BitterLesson}
R.~Sutton.
\newblock The {Bitter} {Lesson}, 2019.
\newblock URL \url{http://www.incompleteideas.net/IncIdeas/BitterLesson.html}.

\bibitem[Tanabe(1971)]{Tanabe1971ProjectionMethodSolving}
K.~Tanabe.
\newblock Projection method for solving a singular system of linear equations and its applications.
\newblock \emph{Numerische Mathematik}, 17\penalty0 (3):\penalty0 203--214, June 1971.
\newblock ISSN 0945-3245.
\newblock \doi{10.1007/BF01436376}.
\newblock URL \url{https://doi.org/10.1007/BF01436376}.

\bibitem[Tay et~al.(2022)Tay, Dehghani, Bahri, and Metzler]{Tay2022EfficientTransformersSurvey}
Y.~Tay, M.~Dehghani, D.~Bahri, and D.~Metzler.
\newblock Efficient {Transformers}: {A} {Survey}.
\newblock \emph{ACM Comput. Surv.}, 55\penalty0 (6):\penalty0 109:1--109:28, Dec. 2022.
\newblock ISSN 0360-0300.
\newblock \doi{10.1145/3530811}.
\newblock URL \url{https://dl.acm.org/doi/10.1145/3530811}.

\bibitem[Trockman et~al.(2024)Trockman, Harutyunyan, Kolter, Kumar, and Bhojanapalli]{Trockman2024MimeticInitializationHelps}
A.~Trockman, H.~Harutyunyan, J.~Z. Kolter, S.~Kumar, and S.~Bhojanapalli.
\newblock Mimetic {Initialization} {Helps} {State} {Space} {Models} {Learn} to {Recall}, Oct. 2024.
\newblock URL \url{http://arxiv.org/abs/2410.11135}.
\newblock arXiv:2410.11135 [cs].

\bibitem[van~der Westhuizen and Lasenby(2018)]{vanderWesthuizen2018UnreasonableEffectivenessForget}
J.~van~der Westhuizen and J.~Lasenby.
\newblock The unreasonable effectiveness of the forget gate, Sept. 2018.
\newblock URL \url{http://arxiv.org/abs/1804.04849}.
\newblock arXiv:1804.04849 [cs, stat].

\bibitem[Vaswani et~al.(2017)Vaswani, Shazeer, Parmar, Uszkoreit, Jones, Gomez, Kaiser, and Polosukhin]{Vaswani2017AttentionAllYou}
A.~Vaswani, N.~Shazeer, N.~Parmar, J.~Uszkoreit, L.~Jones, A.~N. Gomez, L.~Kaiser, and I.~Polosukhin.
\newblock Attention is {All} you {Need}.
\newblock In \emph{Advances in {Neural} {Information} {Processing} {Systems}}, volume~30. Curran Associates, Inc., 2017.
\newblock URL \url{https://papers.nips.cc/paper_files/paper/2017/hash/3f5ee243547dee91fbd053c1c4a845aa-Abstract.html}.

\bibitem[von Oswald et~al.(2023)von Oswald, Niklasson, Schlegel, Kobayashi, Zucchet, Scherrer, Miller, Sandler, y~Arcas, Vladymyrov, Pascanu, and Sacramento]{vonOswald2023UncoveringMesaoptimizationAlgorithms}
J.~von Oswald, E.~Niklasson, M.~Schlegel, S.~Kobayashi, N.~Zucchet, N.~Scherrer, N.~Miller, M.~Sandler, B.~A. y~Arcas, M.~Vladymyrov, R.~Pascanu, and J.~Sacramento.
\newblock Uncovering mesa-optimization algorithms in {Transformers}.
\newblock \emph{CoRR}, abs/2309.05858, 2023.
\newblock URL \url{https://doi.org/10.48550/arXiv.2309.05858}.
\newblock tex.cdate: 1672531200000 tex.publtype: informal.

\bibitem[Wang et~al.(2019)Wang, Pleiss, Gardner, Tyree, Weinberger, and Wilson]{Wang2019ExactGaussianProcesses}
K.~A. Wang, G.~Pleiss, J.~Gardner, S.~Tyree, K.~Q. Weinberger, and A.~G. Wilson.
\newblock Exact {Gaussian} {Processes} on a {Million} {Data} {Points}.
\newblock \emph{Advances in Neural Information Processing Systems}, 32, 2019.
\newblock URL \url{https://papers.nips.cc/paper/2019/hash/01ce84968c6969bdd5d51c5eeaa3946a-Abstract.html}.

\bibitem[Wasserman(2006)]{Wasserman2006AllNonparametricStatistics}
L.~Wasserman.
\newblock \emph{All of {Nonparametric} {Statistics}}.
\newblock Springer {Texts} in {Statistics}. Springer, New York, NY, 2006.
\newblock ISBN 978-0-387-25145-5.
\newblock \doi{10.1007/0-387-30623-4}.
\newblock URL \url{http://link.springer.com/10.1007/0-387-30623-4}.

\bibitem[Widrow and Hoff(1988)]{Widrow1988AdaptiveSwitchingCircuits}
B.~Widrow and M.~E. Hoff.
\newblock Adaptive switching circuits.
\newblock In \emph{Neurocomputing: foundations of research}, pages 123--134. MIT Press, Cambridge, MA, USA, Jan. 1988.
\newblock ISBN 978-0-262-01097-9.

\bibitem[Willshaw(1989)]{Willshaw1989HolographyAssociativeMemory}
D.~Willshaw.
\newblock Holography, {Associative} {Memory}, and {Inductive} {Generalization}.
\newblock In \emph{Parallel {Models} of {Associative} {Memory}}. Psychology Press, 1989.
\newblock ISBN 978-1-315-80799-7.
\newblock Num Pages: 22.

\bibitem[Willshaw et~al.(1969)Willshaw, Buneman, and Longuet-Higgins]{Willshaw1969NonHolographicAssociativeMemory}
D.~J. Willshaw, O.~P. Buneman, and H.~C. Longuet-Higgins.
\newblock Non-{Holographic} {Associative} {Memory}.
\newblock \emph{Nature}, 222\penalty0 (5197):\penalty0 960--962, June 1969.
\newblock ISSN 1476-4687.
\newblock \doi{10.1038/222960a0}.
\newblock URL \url{https://www.nature.com/articles/222960a0}.
\newblock Publisher: Nature Publishing Group.

\bibitem[Wortsman et~al.(2023)Wortsman, Liu, Xiao, Everett, Alemi, Adlam, Co-Reyes, Gur, Kumar, Novak, Pennington, Sohl-Dickstein, Xu, Lee, Gilmer, and Kornblith]{Wortsman2023SmallscaleProxiesLargescale}
M.~Wortsman, P.~J. Liu, L.~Xiao, K.~E. Everett, A.~A. Alemi, B.~Adlam, J.~D. Co-Reyes, I.~Gur, A.~Kumar, R.~Novak, J.~Pennington, J.~Sohl-Dickstein, K.~Xu, J.~Lee, J.~Gilmer, and S.~Kornblith.
\newblock Small-scale proxies for large-scale {Transformer} training instabilities.
\newblock In \emph{The {Twelfth} {International} {Conference} on {Learning} {Representations}}, Oct. 2023.
\newblock URL \url{https://openreview.net/forum?id=d8w0pmvXbZ}.

\bibitem[Xu et~al.(2024)Xu, Cheng, Wang, and Chen]{Xu2024KVShiftingAttention}
M.~Xu, W.~Cheng, B.~Wang, and W.~Chen.
\newblock {KV} {Shifting} {Attention} {Enhances} {Language} {Modeling}, Dec. 2024.
\newblock URL \url{http://arxiv.org/abs/2411.19574}.
\newblock arXiv:2411.19574 [cs].

\bibitem[Yang et~al.(2024{\natexlab{a}})Yang, Kautz, and Hatamizadeh]{Yang2024GatedDeltaNetworks}
S.~Yang, J.~Kautz, and A.~Hatamizadeh.
\newblock Gated {Delta} {Networks}: {Improving} {Mamba2} with {Delta} {Rule}, Dec. 2024{\natexlab{a}}.
\newblock URL \url{http://arxiv.org/abs/2412.06464}.
\newblock arXiv:2412.06464 [cs] version: 1.

\bibitem[Yang et~al.(2024{\natexlab{b}})Yang, Wang, Shen, Panda, and Kim]{Yang2024GatedLinearAttention}
S.~Yang, B.~Wang, Y.~Shen, R.~Panda, and Y.~Kim.
\newblock Gated {Linear} {Attention} {Transformers} with {Hardware}-{Efficient} {Training}.
\newblock In \emph{Proceedings of the 41st {International} {Conference} on {Machine} {Learning}}, pages 56501--56523. PMLR, July 2024{\natexlab{b}}.
\newblock URL \url{https://proceedings.mlr.press/v235/yang24ab.html}.
\newblock ISSN: 2640-3498.

\bibitem[Yang et~al.(2024{\natexlab{c}})Yang, Wang, Zhang, Shen, and Kim]{Yang2024ParallelizingLinearTransformers}
S.~Yang, B.~Wang, Y.~Zhang, Y.~Shen, and Y.~Kim.
\newblock Parallelizing {Linear} {Transformers} with the {Delta} {Rule} over {Sequence} {Length}, Aug. 2024{\natexlab{c}}.
\newblock URL \url{http://arxiv.org/abs/2406.06484}.
\newblock arXiv:2406.06484.

\bibitem[Zhang et~al.(2025)Zhang, Nolte, Sadhukhan, Chen, and Bottou]{Zhang2025MemoryMosaics}
J.~Zhang, N.~Nolte, R.~Sadhukhan, B.~Chen, and L.~Bottou.
\newblock Memory mosaics.
\newblock In \emph{The thirteenth international conference on learning representations}, 2025.
\newblock URL \url{https://openreview.net/forum?id=IiagjrJNwF}.

\bibitem[Zhang et~al.(2023{\natexlab{a}})Zhang, Bhatia, Kumbong, and Re]{Zhang2023HedgehogPorcupineExpressive}
M.~Zhang, K.~Bhatia, H.~Kumbong, and C.~Re.
\newblock The {Hedgehog} \& the {Porcupine}: {Expressive} {Linear} {Attentions} with {Softmax} {Mimicry}.
\newblock In \emph{The {Twelfth} {International} {Conference} on {Learning} {Representations}}, Oct. 2023{\natexlab{a}}.
\newblock URL \url{https://openreview.net/forum?id=4g02l2N2Nx}.

\bibitem[Zhang et~al.(2023{\natexlab{b}})Zhang, Zhang, Yang, and Wang]{Zhang2023WhatHowDoes}
Y.~Zhang, F.~Zhang, Z.~Yang, and Z.~Wang.
\newblock What and {How} does {In}-{Context} {Learning} {Learn}? {Bayesian} {Model} {Averaging}, {Parameterization}, and {Generalization}, Oct. 2023{\natexlab{b}}.
\newblock URL \url{http://arxiv.org/abs/2305.19420}.
\newblock arXiv:2305.19420 [stat].

\bibitem[Zhang et~al.(2024)Zhang, Liu, Cai, Wang, and Wang]{Zhang2024AnalysisAttentionLens}
Y.~Zhang, B.~Liu, Q.~Cai, L.~Wang, and Z.~Wang.
\newblock An {Analysis} of {Attention} via the {Lens} of {Exchangeability} and {Latent} {Variable} {Models}, Apr. 2024.
\newblock URL \url{http://arxiv.org/abs/2212.14852}.
\newblock arXiv:2212.14852 [cs].

\bibitem[Zhou and Troyanskaya(2015)]{Zhou2015PredictingEffectsNoncoding}
J.~Zhou and O.~G. Troyanskaya.
\newblock Predicting effects of noncoding variants with deep learning–based sequence model.
\newblock \emph{Nature Methods}, 12\penalty0 (10):\penalty0 931--934, Oct. 2015.
\newblock ISSN 1548-7105.
\newblock \doi{10.1038/nmeth.3547}.
\newblock URL \url{https://www.nature.com/articles/nmeth.3547}.
\newblock Publisher: Nature Publishing Group.

\bibitem[Zinkevich(2003)]{Zinkevich2003OnlineConvexProgramming}
M.~Zinkevich.
\newblock Online convex programming and generalized infinitesimal gradient ascent.
\newblock In \emph{Proceedings of the {Twentieth} {International} {Conference} on {International} {Conference} on {Machine} {Learning}}, {ICML}'03, pages 928--935, Washington, DC, USA, Aug. 2003. AAAI Press.
\newblock ISBN 978-1-57735-189-4.

\end{thebibliography}
}

\appendix
\appendixpage
\section{Bounding the norm of a linear regression layer}\label{sec:linear-regression-bound}
Let $\ybf_t = \Vbf_t\transpose\Kbf_t(\Kbf_t\transpose\Kbf_t)\inverse \qbf_t$ and let $\norm{\Abf}_2$ indicate the spectral norm of a matrix $\Abf$, a matrix norm induced by the L2 norm on vectors.
Then
\begin{align}
    \norm{\ybf_t}_2 &= \norm{\Vbf_t\transpose\Kbf_t(\Kbf_t\transpose\Kbf_t)\inverse \qbf_t}_2\\
    &\leq \norm{\qbf_t}_2\sup_{\norm{\qbf}_2=1} \norm{\Vbf_t\transpose \Kbf_t (\Kbf_t\transpose\Kbf_t)\inverse\qbf}_2 \\
    & = \norm{\qbf_t}_2\norm{\Vbf_t\transpose \Kbf_t (\Kbf_t\transpose\Kbf_t)\inverse}_2 \\
    &\leq \norm{\qbf_t}_2\norm{\Vbf_t\transpose\Kbf_t}_2\norm{(\Kbf_t\transpose\Kbf_t)\inverse}_2 \\
    & = \norm{\qbf_t}_2\norm{\sum_{i=1}^t \vbf_i\kbf_i\transpose}_2\norm{(\Kbf_t\transpose\Kbf_t)\inverse}_2 \\
    &\leq \norm{\qbf_t}_2\left(\sum_{i=1}^t\norm{ \vbf_i}_2\norm{\kbf_i}_2\right)\norm{(\Kbf_t\transpose\Kbf_t)\inverse}_2 \\
    & = \norm{\qbf_t}_2\left(\sum_{i=1}^t\norm{ \vbf_i}_2\norm{\kbf_i}_2\right)\max_i \lambda_i((\Kbf_t\transpose\Kbf_t)\inverse) \\
    & = \norm{\qbf_t}_2\left(\sum_{i=1}^t\norm{ \vbf_i}_2\norm{\kbf_i}_2\right)\max_i \frac{1}{\lambda_i(\Kbf_t\transpose\Kbf_t)} \\
    & = \frac{\norm{\qbf_t}_2\sum_{i=1}^t\norm{ \vbf_i}_2\norm{\kbf_i}_2}{\lambda_{\text{min}}(\Kbf_t\transpose\Kbf_t)}.
\end{align}
Hence $\norm{\ybf_t}_2 \leq \norm{\qbf_t}_2\sum_{i=1}^t\norm{ \vbf_i}_2\norm{\kbf_i}_2/\lambda_{\text{min}}(\Kbf_t\transpose\Kbf_t)$.
When we approximate $\Kbf_t\transpose\Kbf_t \approx \Ibf$, as in linear attention, we lose the self-normalizing property of dividing by ${\lambda_{\text{min}}(\Kbf_t\transpose\Kbf_t)}$ since the denominator becomes 1.
This explains how output normalization by \citet{Qin2022DevilLinearTransformer} is an attempt at restoring this intrinsic self-normalizing property of linear regression.

\end{document}